\documentclass[10pt,oneside,onecolumn,a4paper]{article}

\usepackage[utf8]{inputenc}
\usepackage[english]{babel}
\usepackage{bibgerm}
\usepackage{verbatim}
\usepackage{graphicx}
\usepackage{hyperref}
\usepackage{tikz}
\usetikzlibrary{arrows,shapes,shadows,positioning}
\newcommand{\copar}[5]{\begin{small}\begin{quote}\underline{{\bf #1}}\end{quote}\begin{list}{}{\setlength{\parsep}{-0.05mm} \setlength{\itemsep}{-0.05mm} \setlength{\leftmargin}{4mm}}\item {\bf description:} #2 \item {\bf type:} #3 \item {\bf values, default:} #4 \item {\bf remarks:} #5 \end{list} \end{small}\hrule}

\newcommand{\codat}[3]{\begin{small}\begin{quote}\underline{{\bf #1}}\end{quote}\begin{list}{}{\setlength{\parsep}{-0.05mm} \setlength{\itemsep}{-0.05mm} \setlength{\leftmargin}{4mm}}\item {\bf description:} #2 \item {\bf type:} #3 \end{list} \end{small}\hrule}

\newcommand{\copac}[1]{\begin{tt}#1\end{tt}}
\newcommand{\mybndfloat}[0]{float $]0$ $100[$}

\newcommand{\spar}[4]{\begin{small}\begin{list}{}{\setlength{\parsep}{-0.05mm} \setlength{\itemsep}{-0.05mm} \setlength{\leftmargin}{4mm}}\item {\bf navigation:} {\em #1} \item {\bf feature sets:} #2 \item {\bf option sub-dictionary:} {\em #3} \item {\bf output sub-dictionary:} {\em #4} \end{list} \end{small}}

\newcommand{\myRef}[1]{\begin{tt}#1\end{tt}}

\title{CoPaSul Manual\\Contour-based, parametric, and superpositional intonation stylization}

\author{Uwe D. Reichel\\Research Institute for Linguistics\\Hungarian Academy of Sciences\\uwe.reichel@nytud.mta.hu}
\date{Version $\geq$ 1.3.0, November 18th, 2023\\\vspace*{0.2cm}}

\begin{document}
\maketitle

\vspace*{2cm}

\begin{center}
  \includegraphics[width=18cm]{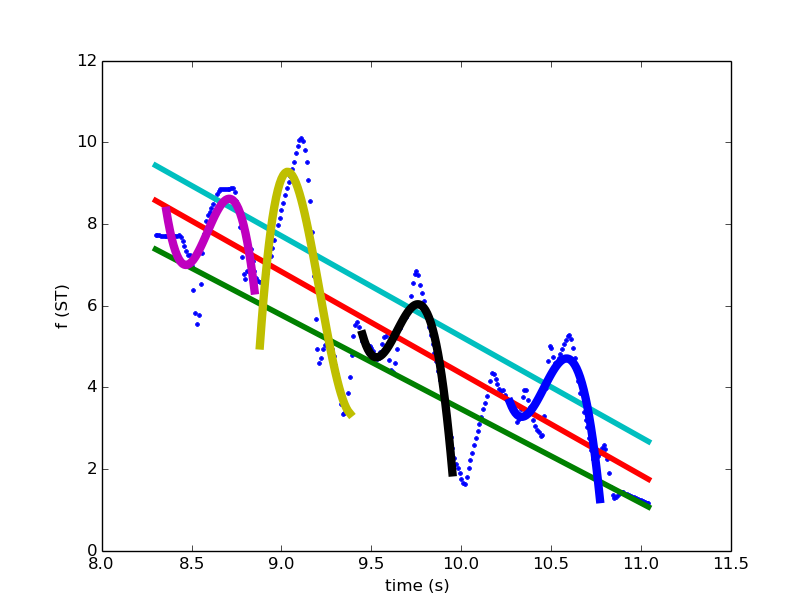}
\end{center}

\newpage

\tableofcontents

\newpage

\section{Introduction}

The purposes of the CoPaSul toolkit are (1) automatic prosodic annotation
and (2) prosodic feature extraction from syllable to utterance level.

CoPaSul stands for {\em contour-based, parametric, superpositional}
intonation stylization. The core model is introduced amongst others in
\cite{ReichelCSL2014}. In this framework intonation is represented as a
superposition of global and local contours that are described
parametrically in terms of polynomial coefficients. On the global
level (usually associated but not necessarily restricted to intonation
phrases) the stylization serves to represent register in terms of
time-varying f0 level and range. On the local level (e.g. accent
groups), local contour shapes are described. From this
parameterization several features related to prosodic boundaries and
prominence can be derived. Furthermore, by coefficient clustering
prosodic contour classes can be derived in a bottom-up way.  Next to
the stylization-based feature extraction also standard f0 and energy
measures (e.g. mean and variance) as well as rhythmic aspects can be
calculated.

At the current state {\bf automatic annotation} comprises:

\begin{itemize}
\item segmentation into interpausal chunks
\item syllable nucleus extraction
\item unsupervised localization of prosodic phrase boundaries and prominent syllables
\end{itemize}

F0 and partly also energy {\bf feature sets} can be extracted for:

\begin{itemize}
\item standard measurements (as median and IQR)
\item register in terms of f0 level and range
\item prosodic boundaries
\item local contour shapes
\item bottom-up derived contour classes
\item Gestalt of accent groups in terms of their deviation from higher
  level prosodic units
\item rhythmic aspects quantifying the relation between f0 and energy
  contours and prosodic event rates
\end{itemize}

Please see section \ref{sec:feat} for a list of application examples.

\section{Download and installation}

\subsection{From GitHub}

The CoPaSul Python3 code can be cloned or downloaded from this location:

\begin{quote}
  \url{https://github.com/reichelu/copasul}
\end{quote}

The software is tested only for Linux but might work on Windows.

\paragraph{The installation steps are:}

\begin{enumerate}
\item change to the copasul/ project folder
\item setup a virtual environment "venv\_copasul", activate it, and
  install the requirements. For Linux this works as e.g. follows:
  \begin{verbatim}
    $ virtualenv --python="/usr/bin/python3" venv_copasul
    $ source venv_copasul/bin/activate
    (venv_copasul) $ pip install -r requirements.txt
  \end{verbatim}
\end{enumerate}

The target directory now contains the subfolders {\em copasul/} and
{\em scripts} with several Python and Praat scripts \cite{Boersma1999}
and a {\em tests/minex/} subfolder with data for a minimal example:

\vspace*{0.2cm}
\centerline{
  \begin{tabular}{l|l}
    \multicolumn{2}{l}{copasul/: Software Python} \\
    \hline
    copasul.py & the main script \\
    copasul\_augment.py & automatic prosodic annotation \\
    copasul\_clst.py & contour clustering \\
    copasul\_export.py & table output \\
    copasul\_init.py & initializations \\
    copasul\_plot.py & stylization plots \\
    copasul\_preproc.py & data preprocessing \\
    copasul\_resyn.py & f0 generation \\
    copasul\_sigproc.py & signal processing \\
    copasul\_styl.py & stylization \\
    copasul\_utils.py & general function collection \\
    \hline
    \multicolumn{2}{l}{scripts/: Software Python, Praat} \\
    \hline
    extract\_f0.praat & f0 extraction for mono files \\
    extract\_f0\_stereo.praat & f0 extraction for stereo files \\
    extract\_pulse.praat & pulse extraction for mono files \\
    extract\_pulse\_stereo.praat & pulse extraction for stereo files \\
    run\_copasul.py & run CoPaSul feature extractor from terminal \\
    \hline
    \multicolumn{2}{l}{docs/: Documentation} \\
    \hline
    copasul\_commented\_config.json.txt & configuration explanations \\
    copasul\_manual\_latest.pdf & latest version of this document \\
    \hline
    \multicolumn{2}{l}{tests/minex/: minimal example} \\
  \end{tabular}
}

\subsection{From PyPi}

CoPaSul is also available as a Python package {\em copasul} and can be
pip-installed with its dependencies as follows:

\begin{verbatim}
$ virtualenv --python="/usr/bin/python3" venv_copasul
$ source venv_copasul/bin/activate
(venv_copasul) $ pip install copasul
\end{verbatim}

\section{Usage}

\subsection{Call from terminal (if cloned from GitHub)}

\begin{verbatim}
(venv_copasul) $ cd scripts/
(venv_copasul) $ python run_copasul.py -c ../tests/minex/config/minex.json
\end{verbatim}

Input and output of this minimal example can be found in the subfolder
{\em tests/minex/}. The content of the configuration file is explained in
section \ref{sec:config}.

\subsection{Example integration into python code}

\begin{verbatim}
import json
import pickle
import sys

# add this line if you use the cloned code from GitHub
# sys.path.append(PROJECT_DIR)

from copasul import copasul

# feature extractor
fex = copasul.Copasul()

# processing based on config file
config_file = MYCONFIGFILE.json
copa = fex.process(config=config_file)

# processing based on config dict
with open(config_file, 'r') as h:
    config_dict = json.load(h)
copa = fex.process(config=config_dict)

# warm start: continue processing
#   - implicit loading from config
config_dict["navigate"]["from_scratch"] = False
config_dict["navigate"]["overwrite_config"] = True
#     ... change further navigation values to your needs
copa = fex.process(config=config_dict)

#   - explicit loading from file
copa_file = MYCOPASULOUTPUT.pickle
config_dict["navigate"]["overwrite_config"] = True
#     ... change further navigation values to your needs
with open(copa_file, "rb") as h:
   copa = pickle.load(h)
copa_further_processed = fex.process(config=config_dict, copa=copa)
\end{verbatim}

The input argument for {\em fex.process()} is a dictionary (or the
name of a json file storing this dictionary) with the configurations
(see section \ref{sec:config}). {\em fex.process()} returns the output
dictionary {\em copa} with the extracted features (see section
\ref{sec:dict}). The feature tabels are stored as Pandas Dataframes
with alphanumerically sorted columns and can be accessed as follows:

\begin{verbatim}
>>> copa['export']['loc']
\end{verbatim}

{\em loc} refers to local contour parameters. Other feature sets are
{\em glob} (global contour parameters), {\em gnl\_f0} (standard f0
features), {\em gnl\_en} (standard energy features), {\em rhy\_f0} (f0
rhythm features) {\em rhy\_en} (energy rhythm features) {\em bnd}
(boundary features), and {\em voi} voice quality features. All
standard and rhythm features are additionally calculated on the file
level ({\em gnl\_f0\_file \ldots rhy\_en\_file}).

For terminal calls as well as for calls within the Python environment
the stylization output is written to a binary Pickle file and to CSV
table files as specified in the configurations. See section
\ref{sec:output}.

\section{Input}

For automatic annotation CoPaSul needs audio and f0 table files. For
feature extraction it additionally needs annotation files. For the
{\em voice} feature set furthermore pulse table files are needed.
Corresponding files do not necessarily need to have the same name
stem, but it is assumed that all audio, f0, and annotation files are
sorted the same. An example can be found in the {\em input\/}
subdirectory.

Additionally a configuration file in {\em JSON} format is needed as
further specified in section \ref{sec:config}.

\subsection{Audio files}

Currently only wav files are supported. The files can be mono or
stereo.  For conversion to wav, e.g. {\em Praat, Audacity}, or {\em
  Sox} software can be used.

\subsection{F0 files}
\label{sec:in_f0}

Plain text files. Tables with whitespace column separator. The first
column contains time information. All further columns contain the f0
of the respective channel. For mono files f0 tables thus consist of 2
columns, for stereo files of 3, etc. All columns need to have the same
lengths. Undefined f0 values are to be replaced by 0. Only 100 Hz
sample rate is supported, and resampling is carried out from other
rates. The Praat scripts {\em extract\_f0.praat} and {\em
  extract\_f0\_stereo.praat} which are contained in this package
provide the required input format.

\subsection{Pulse files}
\label{sec:in_pulse}
Plain text files. Only needed for the {\em voice} feature
extraction. Tables with whitespace column separator. Each column
contains the pulse time stamps for one channel in seconds. All columns
must contain the same number of rows so that for files with more than
one channel -1 has to be padded to the shorter columns. The Praat
scripts {\em extract\_pulse.praat} and {\em extract\_pulse\_stereo.praat}
which are contained in this package provide the required input format.

\subsection{Annotation files}
\label{sec:in_annot}

The Praat TextGrid format (long and short) and an XML format of the
following form are supported.

\begin{small}
\begin{verbatim}
<?xml version="1.0" encoding="UTF-8"?>
<annotation>
  ...
  <tiers>
    <tier>
      <name>mySegmentTier</name>
      <items>
        <item>
          <label>x</label>
          <t_start>0.3</t_start>
          <t_end>0.9</t_end>
        </item>
        ...
      </items>
    </tier>
    <tier>
      <name>myEventTier</name>
      <items>
        <item>
          <label>y</label>
          <t>0.7</t>
        </item>
        ...
      </items>
    </tier>
    ...
  </tiers>
  ...
</annotation>
\end{verbatim}
\end{small}

The tiers need to be stored in the {\em tiers} subtree right below the
root element.

Each tier must have a name assigned by the element {\em name}. The
items of each tier are collected in the {\em items} subtree, in which
each item is stored in an {\em item} subtree.

Segment tiers (see next section) must contain the elements {\em label,
  t\_start, t\_end}. Event tiers must contain the elements {\em label,
  t}.

The XML annotation file can be extended by the user as long as it
fulfills the specified requirements in the {\em tiers} subtree.

\subsection{Annotation tiers}
\label{sec:tiers}

In the following the notation \myRef{a:b:\ldots} refers to branches
through the configuration dictionary which is introduced in section
\ref{sec:config}. The annotation files can contain tiers of the
following types:

\paragraph{Segment tiers}
contain items defined by a label, a start point and an endpoint. They
correspond to Praat IntervalTiers.

\paragraph{Event tiers}
contain items without a temporal extension. They are defined by a
label and a time stamp and correspond to Praat TextTiers.

Both segment tiers and event tiers are supported for most of the
analyses. Wherever needed, an event is converted to a segment by
centering a window of length \myRef{preproc:point\_win} on the event
as is explained in more detail in section \ref{sec:win}. Pause
information can only be extracted for segment tiers. In TextGrids
pauses are considered to be items with empty labels or labeled as
\myRef{fsys:label:pau}. Both event and segment tiers can serve as:

\paragraph{Analysis tiers}
In the context of automatic annotation these tiers contain or limit
the candidate locations for prosodic events. Can be segment or event
tiers.

\begin{quote}
  \myRef{fsys:augment:glob:tier} \\
  \myRef{fsys:augment:loc:tier\_acc} \\
  \myRef{fsys:augment:loc:tier\_ag}
\end{quote}

For feature extraction these segment or event tiers define the units
of analysis.

\begin{quote}
  \myRef{fsys:chunk:tier} \\
  \myRef{fsys:glob:tier} \\
  \myRef{fsys:loc:tier\_acc} \\
  \myRef{fsys:loc:tier\_ag} \\
  \myRef{fsys:bnd:tier} \\
  \myRef{fsys:gnl\_f0:tier} \\
  \myRef{fsys:gnl\_en:tier} \\
  \myRef{fsys:rhy\_f0:tier} \\
  \myRef{fsys:rhy\_en:tier}
\end{quote}

\paragraph{Parent tiers}

Parent tiers (1) limit the analysis and normalization windows by their
segment boundaries. As an example, normalization across chunk
boundaries can be suppressed. (2) They limit the domain of global
trends against which local deviation is measured.  It's strongly
recommended to use segment tiers for this purpose.  If not specified,
the whole file is treated as a single parenting segment. For automatic
annotation parent tiers are to be defined by:

\begin{quote}
  \myRef{fsys:augment:syl:tier\_parent} \\
  \myRef{fsys:augment:glob:tier\_parent} \\
  \myRef{fsys:augment:loc:tier\_parent} \\
\end{quote}

For {\em glob, bnd, gnl\_en, gnl\_f0, rhy\_en, rhy\_f0} feature extraction
(see section \ref{sec:feat}) only speech chunks can serve as parent
domains:

\begin{quote}
  \myRef{fsys:chunk:tier}
\end{quote}

Fallback is again the entire file. For {\em loc} feature extraction
only the segments of the {\em glob} analysis tier can form the parent
domain due to the:

\paragraph{Superpositional framework}

Within the CoPaSul approach (see section \ref{sec:superpos}) the
intonation contour is considered as a superposition of a global and
local components. Their domains are defined by the {\em glob} and {\em
  loc} option branches, respectively:

\begin{quote}
  \myRef{fsys:glob:tier} \\
  \myRef{fsys:loc:tier\_acc} \\
  \myRef{fsys:loc:tier\_ag}
\end{quote}

This has two implications on the annotation tier definitions:

\begin{itemize}
\item for each channel only one tier is supported each for the global and the local local domain
\item the global domain tier is treated as the parent tier for the local domain tier
\end{itemize}

\paragraph{Output tiers}

For automatic annotation these tiers are defined by a stem which
is always expanded by the recording channel index.

\begin{quote}
  \myRef{fsys:augment:chunk:tier\_out\_stm} \\
  \myRef{fsys:augment:syl:tier\_out\_stm} \\
  \myRef{fsys:augment:glob:tier\_out\_stm} \\
  \myRef{fsys:augment:loc:tier\_out\_stm}
\end{quote}

As an example, given a stereo file and the chunk output tier name
CHUNK, the tiers CHUNK\_1 and CHUNK\_2 will be added to the annotation
file. For the sake of an uniform treatment, also for mono files the
channel index will be added.

\paragraph{Tier specification}

For all tiers, that were not automatically generated, the user needs
to specify the recording channel index it refers to (also for mono
files!), e.g.:

\begin{quote}
  \myRef{fsys:channel:'tierA'=1} \\
  \myRef{fsys:channel:'tierB'=2}
\end{quote}

{\em tierA} thus refers to channel 1, and {\em tierB} to channel
2. Tier names can be specified as strings, or as list of strings.

\begin{quote}
  \myRef{fsys:bnd:tier='tierA'}
\end{quote}

means, that the {\em bnd} feature extraction is to be carried out for
units defined by the content of {\em tierA}.

\begin{quote}
  \myRef{fsys:bnd:tier=$[$'tierA','tierB'$]$}
\end{quote}

triggers a {\em bnd} feature extraction for the content of two
tiers. The channels the specified tiers refer to are looked up in
\myRef{fsys:channel:*}.

The name stem of a tier resulting from automatic annotation
(e.g. CHUNK) will be expanded automatically, thus for a chunked stereo
file these two specifications are equivalent:

\begin{quote}
  \myRef{fsys:bnd:tier='CHUNK'}\\
  \myRef{fsys:bnd:tier=$[$'CHUNK\_1', 'CHUNK\_2'$]$}
\end{quote}

For the feature sets {\em bnd, gnl\_en, gnl\_f0, rhy\_en, rhy\_f0}
(see section \ref{sec:feat}) an arbitrary number of tiers can be
specified for each channel. For {\em chunk, glob, loc} only one tier
per channel is supported.

\section{F0 extraction}

For f0 extraction in mono or stereo wav files the two Praat scripts
contained in this package can be used.

They can be called this way:

\begin{verbatim}
> praat extract_f0.praat myStepsize myMinFreq myMaxFreq \
                         myAudioInputDir myF0OutputDir myAudioExt myF0Ext
\end{verbatim}

The usage of extract\_f0\_stereo.praat is the same. Note that
subsequent stylization {\bf in any case initiates a resampling to 100
  Hz}, so that myStepsize here can be directly set to 0.01.  {\em
  myMinFreq} and {\em myMaxFreq} refer to the minimum and maximum of
allowed f0 values in Hz. Values below or above are considered as
measurement errors and are set to 0. The f0 range choice depends on
the recorded speakers.  As a rule of thumb the parameters can be set
to 50 and 400 Hz, respectively. In my {\em myAudioInputDir} the sound
files with the extension {\em myAudioExt} are collected, and
corresponding f0 plain text table files with the audio file's name
stem and the extension {\em myF0Ext} are outputted to the directory
{\em myF0OutputDir}.

\section{Pulse extraction}

Pulse extraction is needed for the {\em voice} feature set only.  For
its extraction in mono or stereo wav files the two Praat scripts
contained in this package can be used.

They can be called this way:

\begin{verbatim}
> praat extract_pulse.praat myMinFreq myMaxFreq \
                         myAudioInputDir myPulseOutputDir myAudioExt myPulseExt
\end{verbatim}

The usage of extract\_pulse\_stereo.praat is the same.  The scripts
make use of Praat's {\em To PointProcess (cc)} routine operating on
sound and pitch objects. For pitch object creation the minimum and
maximum of allowed f0 values {\em myMinFreq} and {\em myMaxFreq} need
to be specified in Hz. In my {\em myAudioInputDir} the sound files
with the extension {\em myAudioExt} are collected, and corresponding
pulse plain text table files with the audio file's name stem and the
extension {\em myPulseExt} are outputted to the directory {\em
  myPulseOutputDir}.

\section{Automatic annotation}
Automatic unsupervised prosodic annotation comprises chunking,
syllable nucleus and boundary extraction, prosodic phrase extraction,
and pitch accent localization. Details of the algorithms will be given
in \cite{ReichelESSV2017}. At the beginning of each introductory
paragraph it is specified:

\spar{which navigation option to set to True in the configuration file
  (see section \ref{sec:config})}{which feature sets result from the
  annotation (see section \ref{sec:feat})}{which configuration
  sub-dictionaries serve to customize the respective processing (see
  section \ref{sec:config})}{which subdirectory of the resulting
  python nested dictionary contains the extracted feature set (see
  section \ref{sec:dict}).}

Paths through the configuration dictionary are referred to by
\myRef{my:path:to:option}.

\spar{do\_augment\_*}{--}{fsys:augment:*:*;
  augment:*:*}{(augmented annotation file)}

\subsection{Chunking}
\label{sec:chunk}

\spar{do\_augment\_chunk}{--}{fsys:augment:chunk:*; augment:chunk:*}{(augmented annotation file)}

Chunking serves to segment the utterance into interpausal units. It is
based on a pause detector, that works the following way: an analysis
window $w_a$ with length \myRef{augment:chunk:l} is moved over the
lowpass-filtered signal together with a longer reference window $w_r$
of length \myRef{augment:chunk:l\_ref} with the same midpoint. A pause
is set where the mean energy in $w_a$ is below a threshold defined
relative to the energy in $w_r$, i.e. if $e(w_a) < e(w_r) \cdot
\textrm{\em augment:chunk:e\_rel}$. Chunks are then trivially assigned
to interpausal intervals. Silence margins can be set at chunk starts
and ends by \myRef{augment:chunk:margin}.

If $w_r$ itself is identified as a pause by $e(w_r) < e(s) \cdot
\textrm{\em augment:chunk:e\_rel}$ it is replaced by $s$; where $s$
consists of selected parts of the acoustic signal in the analysed
channel with absolute amplitude values above the median. By this lower
threshold the robustness against a high occurrence of speech pauses is
increased.

The filtering of the signal can be customized by the sub-dictionary
\myRef{augment:chunk:flt}. In there \myRef{btype} gives the
Butterworth filter type ({\em high, low, band, or none}), \myRef{f}
the cutoff frequencie(s), and \myRef{ord} the order. For pauses as
well as for inter-pause intervals minimum lengths can be defined by
\myRef{augment:min\_pau\_l} and \myRef{min\_chunk\_l},
respectively. Pauses are then merged across too short chunks, and
chunks are merged across too short pauses. The segment tier output
will be added to the annotation file. The tier name is specified by
\myRef{fsys:augment:chunk:tier\_out\_stm} concatenated with the
respective channel index. Standard labels 'x' are assigned to chunk
segments, and \myRef{fsys:label:pau} to the pauses inbetween.

\subsection{Syllable nucleus and boundary extraction}
\label{sec:syl}

\spar{do\_augment\_syl}{--}{fsys:augment:syl:*; augment:syl:*}{(augmented annotation file)}

For syllable nucleus detection the method proposed by
\cite{Pfitzinger1996} is adopted. Again an analysis $w_a$ with length
\myRef{augment:syl:l} and a longer reference window $w_r$ of length
with length \myRef{augment:syl:l\_ref} with the same midpoint are
moved along the signal, which this time is band-pass filtered to focus
on the frequency band related to vocalic nuclei. The filter
specification in \myRef{augment:syl:flt} works as described for
chunking.  From this energy contour the local maxima are extracted. If
for a local maximum the mean energy in $w_a$ supersedes the mean
energy in $w_r$ by a defined factor, i.e. if $e(w_a) > e(w_r) \cdot
\textrm{\em augment:syl:e\_rel}$, and if $e(w_a)$ is not below a
defined fraction of the energy in the current chunk $w_c$ (fallback:
whole file), i.e. $e(w_a) \geq e(w_c) \cdot \textrm{\em
  augment:syl:e\_min}$, a syllable nucleus is set.  From which tier to
get the current chunk is to be defined by
\myRef{augment:syl:tier\_parent}. E.g. it can be the output tier of a
preceding chunking step. A further constraint
\myRef{augment:syl:d\_min} specifies the minimum distance between
subsequent syllable nuclei. If two nuclei are too close, they are
merged to a single syllable and the point of energy maximum in this
interval is assigned to be the nucleus.

Subsequently syllable boundaries are assigned to the energy minimum
between adjacent syllable nuclei. They just serve as fallback prosodic
boundary candidates.

The output consists of two event tiers for syllable nuclei and
boundaries and will be added to the annotation file. The tier name is
specified by \myRef{fsys:augment:syl:tier\_out\_stm}. For the nuclei
it is concatenated with the respective channel index. For the
boundaries it is concatenated with a 'bnd' infix and the channel
index. Standard labels 'x' are assigned for both tiers.

\subsection{Prosodic phrase boundary location}
\label{sec:augbnd}

\spar{do\_augment\_glob}{--}{fsys:augment:glob:*; augment:glob:*}{(augmented annotation file)}

Prosodic phrase boundary decisions are based on nearest centroid
classification. The user needs to specify the tier that contains
boundary candidates in \myRef{fsys:augment:glob:tier}.  For segment
tiers these candidates are the segment boundaries, for event tiers,
the candidates are the time stamps. If no tier is specified, syllable
boundaries derived by step \ref{sec:syl} will be selected as
candidates. At each boundary candidate a feature set is extracted that
had been proven to be related to prosodic boundaries in former studies
\cite{ReichelESSV2013, ReichelMadyIS2014}. This feature set is
introduced in section \ref{sec:bnd}. The user needs to specify which
of these features should be selected by 

\begin{quote}
  \myRef{augment:glob:wgt:myBndFeatset+:myRegister+:myFeat+}.
\end{quote}

In case a phone segment tier is available and if centroids are derived
from the entire data set and not separately for each file (see below),
in addition z-scored vowel length can be used as a feature. The
length of the vowel associated with the prosodic event candidate is
divided by its mean length derived from the entire dataset. The
associated vowel is the last vowel segment with an onset before the
boundary candidate time stamp. The length feature can be added by:

\begin{quote}
  \myRef{augment:glob:wgt:pho=1}
\end{quote}

The phonetic segment tiers (one for each channel) are to be specified in

\begin{quote}
\myRef{fsys:pho:tier}
\end{quote}

Vowels are identified in these tiers by a regular expression stored in

\begin{quote}
\myRef{fsys:pho:vow}
\end{quote}

This feature will be beneficial for languages in which phrase
boundaries and/or accents are marked by phone segment lengthening.

Furthermore the user can select whether the current feature values at
time $i$, $v_i$, or the delta values (i.e. the differences to the
preceding values $v_i-v_{i-1}$) or both should be taken:

\begin{quote}
  \myRef{augment:glob:measure}
\end{quote}

Some features require units from a parent tier which is to be
specified by \myRef{augment:glob:tier\_parent}, e.g. to measure local
f0 trend discontinuities within a superordinate unit and to limit
analysis and normalization windows. Such units are e.g. chunks derived
from preceding chunking. Fallback is the entire file.

From the features for each of the two classes {\em boundary B} and
{\em no boundary NB} a centroid can be bootstrapped in several ways
given the specification in \myRef{augment:glob:cntr\_mtd} as described
in the following sections. Centroids can be calculated separately for
each file or over the entire data set by setting the value of

\begin{quote}
  \myRef{augment:glob:unit}
\end{quote}

to {\em file} or {\em batch}, respectively. The latter is strongly
recommended for corpora containing lots of short recordings.

\subsubsection{Percentile split}

\begin{quote}
  \myRef{augment:glob:cntr\_mtd=split} \\
  \myRef{augment:glob:prct=mySplitPoint}
\end{quote}

Since for all extracted pause length and pitch discontinuity boundary
features are positive correlation has been found to perceived boundary
strength \cite{ReichelESSV2013, ReichelMadyIS2014} B and NB centroids
can be straight-forwardly derived from high and low feature values,
respectively. Centroids are thus derived by splitting each column in
the feature matrix at the percentile \myRef{augment:glob:prct}. The B
centroid is defined by the median of the values above the splitpoint,
the NB centroid by the median of the values below. All feature vectors
are then assigned to the nearest centroid in a single pass. Boundaries
are subsequently inserted at all candidate time points classified as B. This
method works for both segment and event tier input.

\subsubsection{Bootstrapping seed centroids for kMeans}

\begin{quote}
  \myRef{augment:glob:cntr\_mtd=seed\_kmeans} \\
  \myRef{augment:glob:min\_l=myMinPhraseLength}
\end{quote}

\begin{figure}[!ht]
  \centering
  \includegraphics[width=8cm]{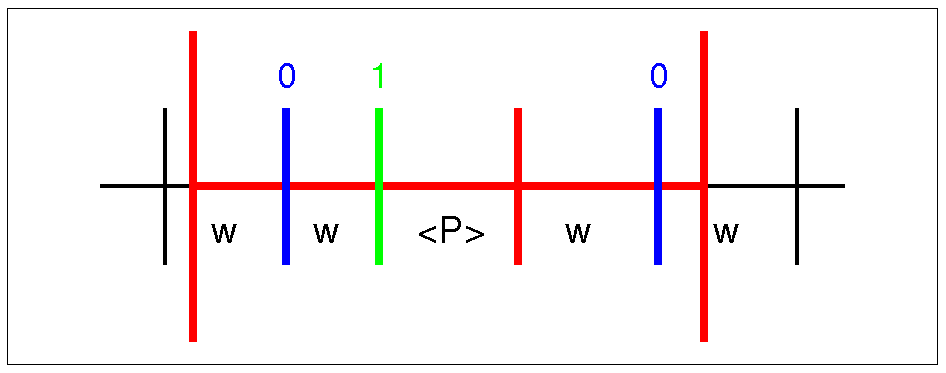}
  \caption{Bootstrapping seed centroids for the classes 1 (boundary)
    and 0 (no boundary). Word boundaries are indicated by the short
    vertical lines. Assumptions: each pause indicates a prosodic
    boundary (green), and prosodic phrases have a minimum length
    (red window), thus in the vicinity of pauses there are no further
    boundaries (blue).}
  \label{fig:boot_bnd}
\end{figure}

This procedure works for segment tier input only since it makes use of
pauses between adjacent segments. As visualized in Figure
\ref{fig:boot_bnd} B and NB centroids are bootstrapped based on two
assumptions: (1) each pause indicates a prosodic boundary, and (2)
prosodic phrases have a minimum length, thus in the vicinity of pauses
there are no further boundaries. KMeans clustering is then initialized
by these two centroids and subdivides all candidates into the B and NB
cluster. Boundaries are inserted at all candidate time points
belonging to the B cluster.

\subsubsection{Bootstrapping seed centroids for percentile split}

\begin{quote}
  \myRef{augment:glob:cntr\_mtd=seed\_prct} \\
  \myRef{augment:glob:prct=mySplitPoint} \\
  \myRef{augment:glob:min\_l=myMinPhraseLength}
\end{quote}

The seed centroid bootstrapping works as for the preceding method.
Instead of kMeans, for the remaining feature vectors the Euclidean
distance to the NB seed centroid is calculated. Vectors with a
distance above the {\em mySplitPoint}-th percentile of all measured
distances are assigned to the B class, the others to the NB class.

\subsubsection{Practical considerations}
The percentile split method works for both segment and event tiers,
whereas the two centroid bootstrapping methods need segment tier input
to infer pause locations. For the two percentile split approaches, the
parameter \myRef{augment:glob:prct} serves to control for the number of
inserted boundaries. The higher, the smaller the B class, thus the
fewer boundaries will be assigned.

If a text transcription is at hand the user can ensure that prosodic
boundaries only occur at word boundaries by preceding signal-text
alignment, e.g. by WebMAUS \cite{Schiel1999, KRSDJP_LREC2016}.

\paragraph{Heuristics}

\begin{quote}
  \myRef{augment:glob:heuristics=ORT}
\end{quote}

If set by the user, this heuristics assumes a word segment tier as
input and rejects boundaries after too short and thus probably
function words ($<0.2s$)

\subsubsection{Feature selection and weighting}
\label{sec:featwgt}

\begin{quote}
 \myRef{augment:glob:wgt:myBndFeatset+:myRegister+:myFeat+=myWeight} \\
 \myRef{augment:glob:wgt\_mtd=myWeightingMethod}
\end{quote}

By the \myRef{augment:glob:wgt:myBndFeatset+:myRegister+:myFeat+}
branches the user at the same time selects and weights features. As an
example

\begin{quote}
  \myRef{augment:glob:wgt:win:ml:rms=1}
\end{quote}

selects the feature \myRef{rms} derived from the register representation
\myRef{ml} within the boundary feature set \myRef{win} (see sections
\ref{sec:bnd} and \ref{sec:feat} for explanations). If the weighting
method in \myRef{augment:glob:wgt\_mtd} is set to 'user', the weight
of this feature becomes 1. If no weighting is intended, to all
selected features should be assigned the same weight. As an
alternative to the definition by the user, weights can also be
extracted by correlation to the median or by the cluster silhouette
measure.

\paragraph{Correlation}
Each feature is correlated with the medians of the feature
vectors. Since as mentioned all boundary features are expected to be
positively correlated to boundary strength, and since the median is
expected to be more robustly related to boundary strength than single
features, the correlation between a feature and the medians to some
extend reflects the goodness of this feature to predict boundary
strength. Features with a negative correlation to the median will be
removed from the pool. All remaining correlations are transformed to
weights summing up to 1 by dividing them by the sum of correlations.

\paragraph{Silhouette}
The mean silhouette over all clustered data points measures how well
clusters can be separated. Here it is measured separately for each
feature within the clearly assignable feature vectors from which the B
and NB seed centroids were derived. It is minmax-normalized to the
range $[0$ $1]$.

\subsubsection{Output}
The output consists of a segment tier for each channel with the name
\myRef{fsys:glob:tier\_out\_stm} + channelIndex. Each segment spans the
interval between two subsequent B events. If \myRef{fsys:glob:tier} is a
segment tier, then pauses are taken over from this tier.  Standard
labels 'x' are assigned to the prosodic phrase segments.

\subsection{Pitch accent detection}
\label{sec:augacc}

Pitch accents are derived in an analogous bootstrap fashion as
prosodic boundaries. The user needs to specify an event tier (default:
syllable nuclei) for localization of the pitch accent
candidates. Furthermore the user can specify a segment tier
(e.g. words) to restrict the maximum number of detected pitch accents
within each segment to 1.

\begin{quote}
  \myRef{fsys:augment:loc:tier\_acc} \\
  \myRef{fsys:augment:loc:tier\_ag}
\end{quote}

Given a segment tier, the user can furthermore specify (1) whether
each segment should get an accent or only the prominent ones

\begin{quote}
  \myRef{augment:loc:ag\_select}
\end{quote}

and (2) where within a segment an accent should be placed: left- or
rightmost, e.g. for prosodically left- or right-headed languages, or
on the most prominent candidate.

\begin{quote}
  \myRef{augment:loc:acc\_select}
\end{quote}

Prominence can be parameterized by several feature sets measuring
standard f0 and energy features, contour shapes within local segments
and their deviation from a global declination trend.

The user can select whether the current feature values at time $i$,
$v_i$, or the delta values (i.e. the differences to the preceding
values $v_i-v_{i-1}$) or both should be taken:

\begin{quote}
  \myRef{augment:loc:measure}
\end{quote}

Some features require units from a parent tier which is to be
specified by \myRef{augment:loc:tier\_parent}, e.g. to measure local
f0 deviations relative to some superordinate unit and to limit
analysis and normalization windows. Such units are e.g. prosodic
phrases derived from preceding phrase extraction. Fallback is the
entire file.

From these features for each of the two classes {\em accented A} and
{\em not accented NA} a centroid can be bootstrapped in several ways
analogously to the prosodic boundary extraction, this time given the
specification in \myRef{augment:loc:cntr\_mtd}.

Centroids can be calculated separately for each file or over the
entire data set by setting the value of

\begin{quote}
  \myRef{augment:loc:unit}
\end{quote}

to {\em file} or {\em batch}, respectively. The latter is strongly
recommended for corpora containing lots of short recordings.

\subsubsection{Percentile split}

\begin{quote}
  \myRef{augment:loc:cntr\_mtd=split} \\
  \myRef{augment:loc:prct=mySplitPoint}
\end{quote}

Given a user-defined feature set where for each feature high values
indicate prominence A and NA centroids can be straight-forwardly
derived from high and low feature values, respectively. Centroids are
thus derived by splitting each column in the feature matrix at the
percentile \myRef{augment:loc:prct}. The A centroid is defined by the
median of the values above the splitpoint, the NA centroid by the
median of the values below. All feature vectors are then assigned to
the nearest centroid in a single pass. Boundaries are then inserted at
all candidate time points classified as B. This method works for both
segment and event tier input.

\subsubsection{Bootstrapping seed centroids for kMeans}

\begin{quote}
  \myRef{augment:loc:cntr\_mtd=seed\_kmeans} \\
  \myRef{augment:loc:max\_l\_na=myMaxLengthNA} \\
  \myRef{augment:loc:min\_l\_a=myMinLengthA} \\
  \myRef{augment:loc:min\_l=myMinLengthAG} 
\end{quote}

\begin{figure}[!ht]
  \centering
  \includegraphics[width=8cm]{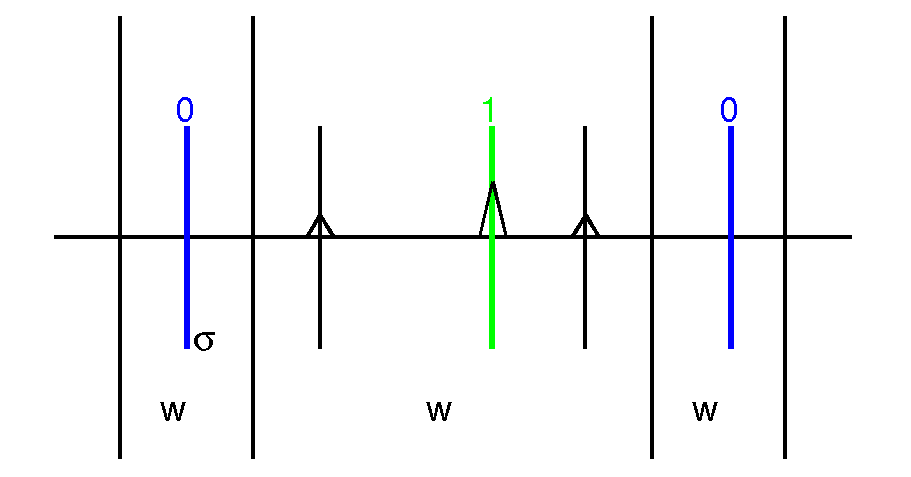}
  \caption{Bootstrapping seed centroids for the classes 1 (accent) and
    0 (no accent). Word boundaries are indicated by long vertical
    lines, and syllable nuclei by short vertical lines. Prominence is
    encoded by the size of the triangles. Assumptions: each word longer
    than some threshold contains an accent (green); each word shorter
    than some threshold does not contain an accent (blue). Within the
    accented word the accent is placed on the most prominent syllable
    (as in this example), or on the left- or rightmost syllable.}
  \label{fig:boot_acc}
\end{figure}

This procedure works only if a segment tier is provided next to the
event tier, and if this segment tier contains word-like units. As for
the phrase boundary detection described above there are 2 (this time
even more) simplifying assumptions to derive seed centroids for
cluster initialization (cf. Figure \ref{fig:boot_acc}): (1) each word
longer than \myRef{augment:loc:min\_l\_a} contains an accent, due to
its expected high information content. (2) each word shorter than
\myRef{augment:loc:max\_l\_na} does not contain an accent due to its
expected low information content. Depending on
\myRef{augment:loc:acc\_select} the A centroid is then calculated from
all leftmost, rightmost, or most prominent \myRef{tier\_acc}
candidates in the \myRef{tier\_ag} segments fulfilling criterion
(1). The NA centroid is calculated from all \myRef{tier\_acc}
candidates in in the \myRef{tier\_ag} segments fulfilling criterion
(2). KMeans clustering is then initialized by these two centroids and
subdivides all candidates into the A and NA cluster. Multiple A cases
within the same segment are reduced by
\myRef{augment:loc:acc\_select}. Furthermore, among A cases closer
than \myRef{augment:loc:min\_l} only the more prominent ones are kept.

\subsubsection{Bootstrapping seed centroids for percentile split}

\begin{quote}
  \myRef{augment:loc:cntr\_mtd=seed\_prct} \\
  \myRef{augment:loc:prct=mySplitPoint} \\
  \myRef{augment:loc:max\_l\_na=myMaxLengthNA} \\
  \myRef{augment:loc:min\_l\_a=myMinLengthA}
\end{quote}

The seed centroid bootstrapping works as for the preceding method.
Instead of kMeans, for the remaining feature vectors the Euclidean
distance to the NA seed centroid is calculated. Vectors with a
distance above the {\em mySplitPoint}-th percentile of all measured
distances are assigned to the A class, the others to the NA class.

\subsubsection{Practical considerations}
The percentile split method works with and without segment tiers,
whereas the two centroid bootstrapping methods need segment tier input
next to the event tier to infer word length. As with boundary
detection, the parameter \myRef{augment:loc:prct} serves to control for
the number of assigned accents. The higher, the smaller the A
class, thus the fewer accents will be assigned.

As mentioned for prosodic boundary detection, a supporting word
segmentation can be derived by preceding signal-text alignment,
e.g. by WebMAUS \cite{Schiel1999, KRSDJP_LREC2016}.

\subsubsection{Feature selection and weighting}

\begin{quote}
  \myRef{augment:loc:wgt:myFeatset+:\ldots}
\end{quote}

The same selection and weighting mechanisms apply as described in
section \ref{sec:featwgt}.

The following feature sets can be used: {\em acc, gst, gnl\_f0,
  gnl\_en} (see section \ref{sec:feat}). In section \ref{sec:config}
examples are given how to expand the corresponding configuration
branches.

As for boundary detection also for pitch accent detection z-scored
vowel length can be added to the feature set. The vowel interval
associated to a pitch accent candidate includes the candidate's time
stamp. See section \ref{sec:augbnd} for further details. The length
feature can be added by:

\begin{quote}
  \myRef{augment:loc:wgt:pho=1}
\end{quote}

\subsubsection{Output}
The output consists of an event tier for each channel with the name
\myRef{fsys:loc:tier\_out\_stm} + channelIndex.  Standard labels 'x'
are assigned to each accent.

\section{Stylization}

In the following the f0 preprocessing and the f0 and energy
stylization steps are introduced. For each stylization step it is specified:

\spar{which navigation option to set to True in the configuration file
  (see section \ref{sec:config})}{which feature sets result from the
  stylization (see section \ref{sec:feat})}{which configuration parts
  serve to customize the respective processing (see section
  \ref{sec:config})}{which part of the resulting Python nested
  dictionary variable contains the extracted feature set (see section
  \ref{sec:dict}).}

Branches through the configuration as well as trough the result
dictionary are referred to by \myRef{my:branch:to:value}.

\subsection{F0 preprocessing}

F0 preprocessing comprises resampling to 100 Hz, outlier detection,
interpolation over outliers and voiceless utterance parts, smoothing,
and semitone conversion including speaker normalization.

\paragraph{Outliers}
Outliers are identified separetely for each channel in a file. They
are defined in terms of deviation from a mean value or from the 1st
and 3rd quartile. The deviation factor is controlled by
\myRef{preproc:out:f}, and the reference point by
\myRef{preproc:out:m}. For \myRef{m=mean} outliers lie outside the
interval $[m-f \cdot \textrm{sd}, m+f \cdot \textrm{sd}]$. For
\myRef{m=median} outliers lie outside of $[m-f \cdot \textrm{iqr}, m+f
  \cdot \textrm{iqr}]$. For \myRef{m=fence} outliers lie outside of
$[Q_1 - f \cdot \textrm{iqr}, Q_3 + f \cdot \textrm{iqr}]$ (sd:
standard deviation; iqr: interquartile range; Q1, Q3: 1st and 3rd
quartile).

\paragraph{Interpolation}
Only linear interpolation is supported. Horizontal extrapolation is
carried out at file boundaries.

\paragraph{Smoothing}
The smoothing method is chosen by \myRef{preproc:smooth:mtd}. Median
and Savitzky-Golay filtering are supported. Median filtering yields
smoother contours, while Savitzky-Golay better preserves local f0
maxima and minima. The higher the window length
\myRef{preproc:smooth:win}, the more smooth the contours. For the
Savitzky-Golay filtering the polynomial order needs to be specified by
\myRef{preproc:smooth:ord}. The lower, the more the result gets
smoothed away from the input data.

\spar{do\_preproc}{--}{preproc:*}{data:myFileIdx:myChannelIdx:f0:*}

\paragraph{Semitone conversion}
If \myRef{preproc:st=1}, Hertz (Hz) values are transformed to
semitones (st) as follows: $F0_{st} = 12 \cdot
log_2(\frac{F0_{Hz}}{b})$. $b$ is a base value which is calculated
separately for each channel in each f0 file. It is defined as the
median of the values below the percentile \myRef{preproc:base\_prct}
and can be used for f0 normalization by file and
channel. Alternatively, a grouping variable can be specified, so that
for each of its levels a separate f0 base value is calculated. This is
done by \myRef{preproc:base\_prct\_grp}. There it can be specified
which grouping variable is to be assigned to each channel. The
grouping variable must be encoded in the filename and must be
extractable from \myRef{fsys:grp:lab}. An example: you have stereo f0
files with the name pattern {\em
  speakerChannel1\_speakerChannel2}. And you want to calculate
separately for each speaker an f0 base value which is the median of
the values below the 5th percentile over all this speaker's utterances
in the corpus. This is to be configured as follows:

\begin{quote}
\myRef{fsys:grp:src='f0'} \\
\myRef{fsys:grp:sep='\_'} \\
\myRef{fsys:grp:lab=$[$'speakerChannel1','speakerChannel2'$]$} \\
\myRef{preproc:base\_prct=5} \\
\myRef{preproc:base\_prct\_grp:'1'='speakerChannel1'} \\
\myRef{preproc:base\_prct\_grp:'2'='speakerChannel2'}
\end{quote}

This assigns to each channel the grouping variable to be read from the
f0 file names. Note, that (1) channel indices need to be written in
quotation marks, and (2) a shared semantics across the grouping
variables is assumed. E.g. just one base value will be calculated for
speaker {\em x}, regardless whether she was recorded in channel 1 or
2.

\paragraph{Base value subtraction}
If \myRef{preproc:st} is $0$, the base value introduced in the
preceding paragraph will be subtracted from the f0 contour without
semitone conversion. If you don't want to use any base value, neither
for subtraction nor as conversion reference, set
\myRef{preproc:base\_prct=0}.

\subsection{Energy calculation}
\label{ssec:en}
The energy contour is simply represented in terms of the root mean
squared deviation (RMSD) within the windowed signal. The relevant
parameters can be found below \myRef{styl:gnl\_en}
\myRef{styl:rhy\_en:sig}. \myRef{win} defines the window length and
\myRef{sts} the stepsize. The energy value sample rate is thus 1/{\em
  sts}. {\em wintyp} and \myRef{winparam} give the window type and an
additional parameters passed on the {\em get\_window()} function of
the {\em scipy.signal} module. For customizing energy extraction with
other than default values, please consult the {\em scipy.signal}
documentation for {\em get\_window()}. {\em wintyp} and {\em winparam}
can contain any value specified in this documentation.

\subsection{Analysis and normalization windows}
\label{sec:win}

Windows serve (1) to transform time stamps from an event tier to
segments, and (2) to locally normalize feature values.

\begin{figure}[!ht]
  \centering
  \includegraphics[width=8cm]{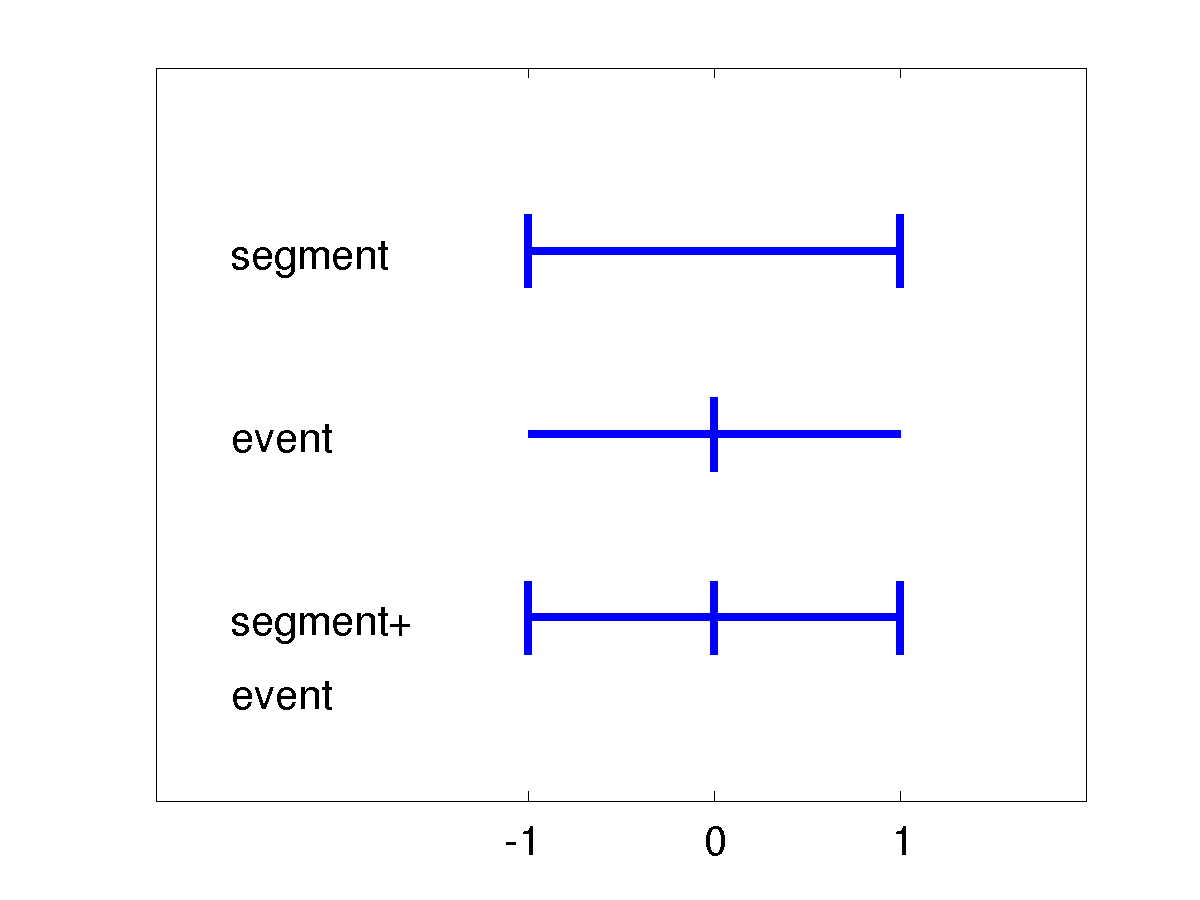}
  \caption{Segment and event tier input. A symmetric analysis window
    is centered on events. For local contour stylization, segment and
    event tiers can be integrated for time normalization: the event is
    set to 0, the pre-event part of the segment to $[-1$ $0[$, and the
        post-event part of the segment to $]0$ $1]$.}
  \label{fig:tierType}
\end{figure}

\paragraph{Time stamps to segments}
Most feature sets are calculated for segments, not for time
stamps. Thus event tier input is converted to segments by centering a
symmetric analysis window with the length \myRef{preproc:point\_win}
on each time stamp as shown in Figure \ref{fig:tierType}. Features are
then extracted within this window. The window can also be separately
specified for each feature set by
\myRef{preproc:myFeatureSet:point\_win}. For local contour stylization
a segment and an event tier can be processed in parallel as explained
in section \ref{sec:loc}.

\paragraph{Normalization}
For the feature sets {\em loc, gnl\_f0} and {\em gnl\_en} several
feature values are additionally locally normalized to capture their
relative amount compared to the local environment. This environment
length is defined by \myRef{preproc:nrm\_win}.  For event tier input
the normalization window is centered on each time stamp. For segment
tier input, it is centered on the midpoint of each segment. For
parallel segment and event tier input which can be provided for {\em
  loc} feature extraction, the window is centered on the event's time
stamp within the segment. The window can also be separately specified
for each feature set by \myRef{preproc:myFeatureSet:nrm\_win}.

\begin{figure}[!ht]
  \centering
  \includegraphics[width=8cm]{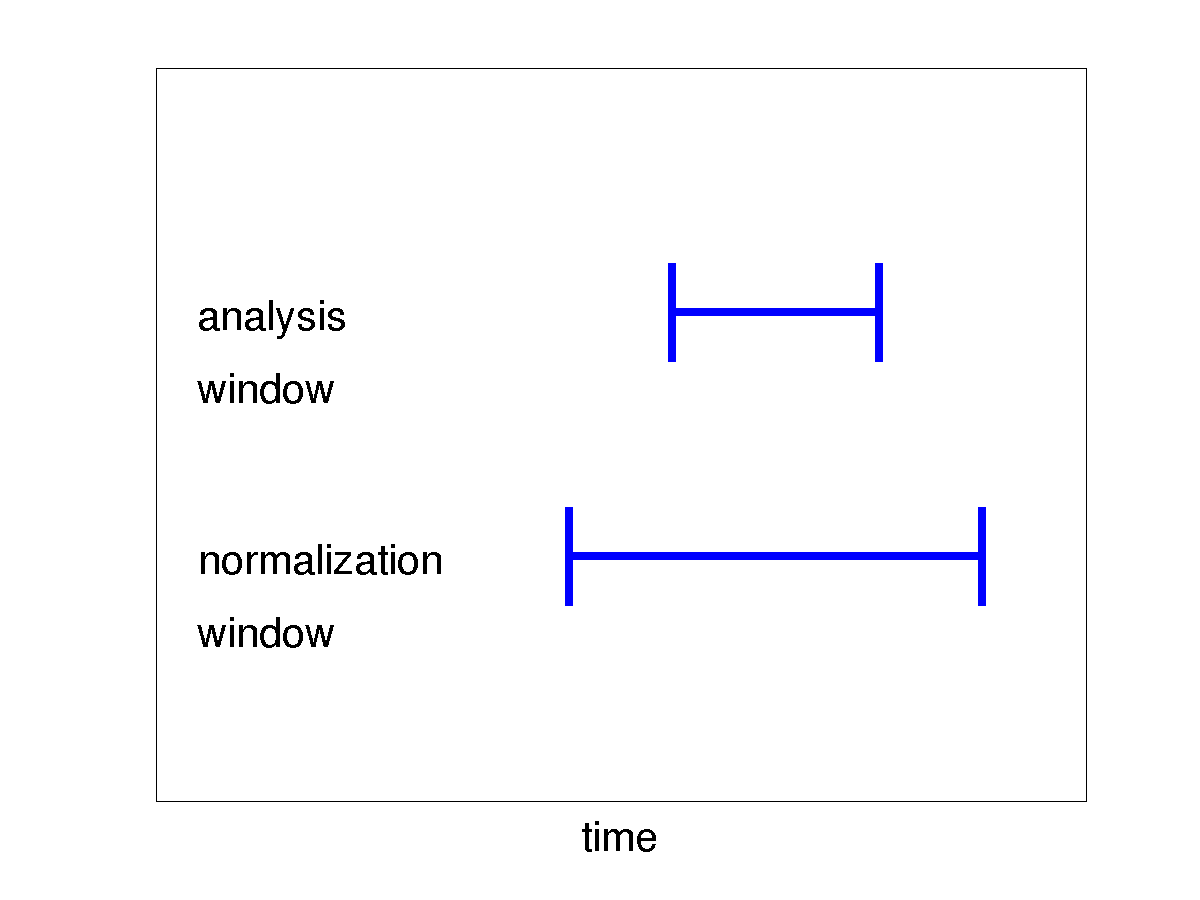}
  \caption{Analysis and longer normalization window. The values
    derived in the analysis window are divided by the corresponding
    values in the normalization window.}
  \label{fig:nrmWin}
\end{figure}

\paragraph{Window constraints}
Analysis and normalization window are limited to the corresponding
segment in the parent tier domain. For {\em loc} features this domain
is given by the global segment tier. For the other features it is
given by the speech chunk tier if this tier is defined in {\em
  fsys:chunk:tier}.  This means that analysis and normalization is not
carried out across global segments or chunks, respectively.  An
exception can be made for the {\em bnd} feature set, that might be
meaningful for chunk boundaries, too. If so,
\myRef{styl:bnd:cross\_chunk} is to be set to 1. For segment tier
input the minimum length of the normalization window is set to the
length of the respective segment. This implies that for segments
longer than the defined normalization window, normalized feature
values are the same as the not normalized ones.

\spar{do\_preproc}{--}{preproc:*}{data:myFileIdx:myChannelIdx:\ldots:$\{t|to|tn\}$}

\subsection{Superposition}
\label{sec:superpos}

\begin{figure}[!ht]
  \centering
  \includegraphics[width=8cm]{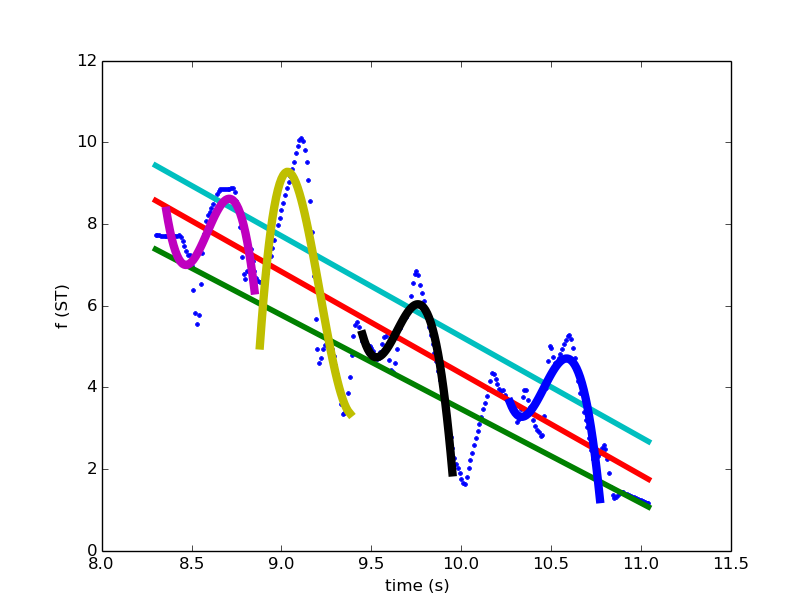}
  \caption{Superposition of one global and four local contours.}
  \label{fig:superpos}
\end{figure}

The core concept of CoPaSul is to represent an f0 contour as a
superposition of linear global component and polynomial local
components as shown in Figure \ref{fig:superpos}.

Stylization is carried out as follows: Within each global segment of
the tier \myRef{fsys:glob:tier} (e.g. an intonation phrase) a linear
register level and range representation is fitted. After subtraction
of this global component, within each local segment an n-th order
polynomial is fitted to the f0 residual. As an alternative to register
level subtraction, the f0 residual can also be derived by
normalization of the contour to the register range.

\subsection{Global segments}
\label{sec:glob}

\subsubsection{Annotation}
\label{sec:globseg_annot}
In the annotation files global segments can be defined in 2 ways:

\begin{enumerate}
\item by start and end point (segment tier input specified in
  \myRef{fsys:glob:tier})
\item by the segments' right end points (event tier input specified in
  \myRef{fsys:glob:tier} that contains e.g. break index labels)
\end{enumerate}

In the second case the events are expanded to segments between the
annotated boundary time stamps. Pauses marked by an empty label or a
pause label (\myRef{fsys:label:pau}) are skipped and the onset of the
subsequent segment is set to the end of the pause. Therefore, in point
tiers pauses should be marked at their right end. Furthermore, if
chunks are provided by \myRef{fsys:tier:chunk}, then the expanded
segments do not cross chunk boundaries but end and start with the
boundaries of the respective chunk they are part of.

\subsubsection{Register}
\label{ssec:register}

Global segments are represented in terms of a time-varying f0 register.
Register aspects are level (midline) and range (topline $-$ baseline).

\begin{figure}[!ht]
  \centering
  \includegraphics[width=8cm]{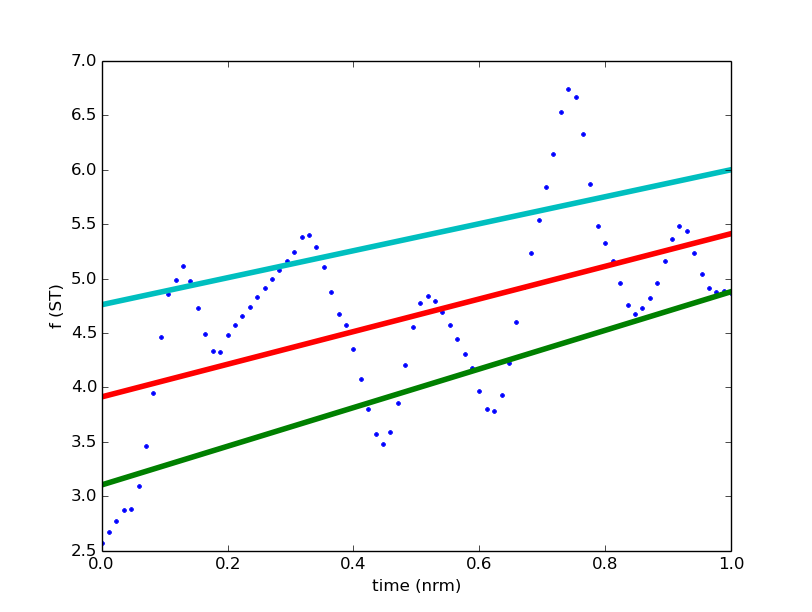}
  \caption{Register (level and range) stylization in global contour
    segments.}
  \label{fig:glob}
\end{figure}

\spar{do\_styl\_glob}{glob}{styl:glob:*}{data:myFileIdx:myChannelIdx:glob:*}

The register fitting procedure consists of the following steps:

\begin{itemize}
\item A window of length \myRef{styl:glob:decl\_win} is shifted along
  the f0 contour with a step size of 10 ms.
\item Within each window the f0 median is calculated
\begin{itemize}
\item of the values below the \myRef{styl:glob:prct:bl} percentile for the baseline,
\item of the values above the \myRef{styl:glob:prct:tl} percentile for
  the topline, and
\item of all values for the midline.
\end{itemize}
This gives 3 sequences of medians, one for the base-, the mid-, and
the topline, respectively.
\item To each of the three median sequences a linear regression line
  is fitted. To be able to compare contours across global segments of
  different lengths, time is normalized as specified by
  \myRef{styl:glob:nrm:mtd} to the range \myRef{styl:glob:nrm:rng}.
\end{itemize}

The motivation for using f0 medians relative to respective percentiles
instead of local peaks and valleys is twofold. First, the stylization
is less affected by prominent pitch accents and boundary
tones. Second, errors resulting from incorrect local peak detection
are circumvented. Both enhances stylization robustness as is shown in
\cite{ReichelMadyIS2014}.

The following configuration parameters serve to customize how closely
the base- and topline should follow local minima and maxima:

\begin{quote}
\myRef{styl:glob:prct:bl}\\
\myRef{styl:glob:prct:tl}\\
\myRef{styl:glob:decl\_win}
\end{quote}

A closer fit to local peaks and valleys is achieved by lowering
\myRef{styl:glob:prct:bl} and \myRef{styl:glob:decl\_win}, and by
raising \myRef{styl:glob:prct:tl}. Note however, that a closer fit
will result in a higher percentage of base- and topline crossings.

From this stylization, regression line slope and intercept features
are collected for the base-, mid-, and topline, as well as for the
range. For the latter these features are simply derived by fitting a
linear regression line through the point-wise distances between the
base- and the topline. A negative slope means that base- and topline
converge, whereas a positive slope signals line divergence.

\subsubsection{Contour classes}
\label{sec:clst_glb}

\begin{figure}[!ht]
  \centering
  \includegraphics[width=8cm]{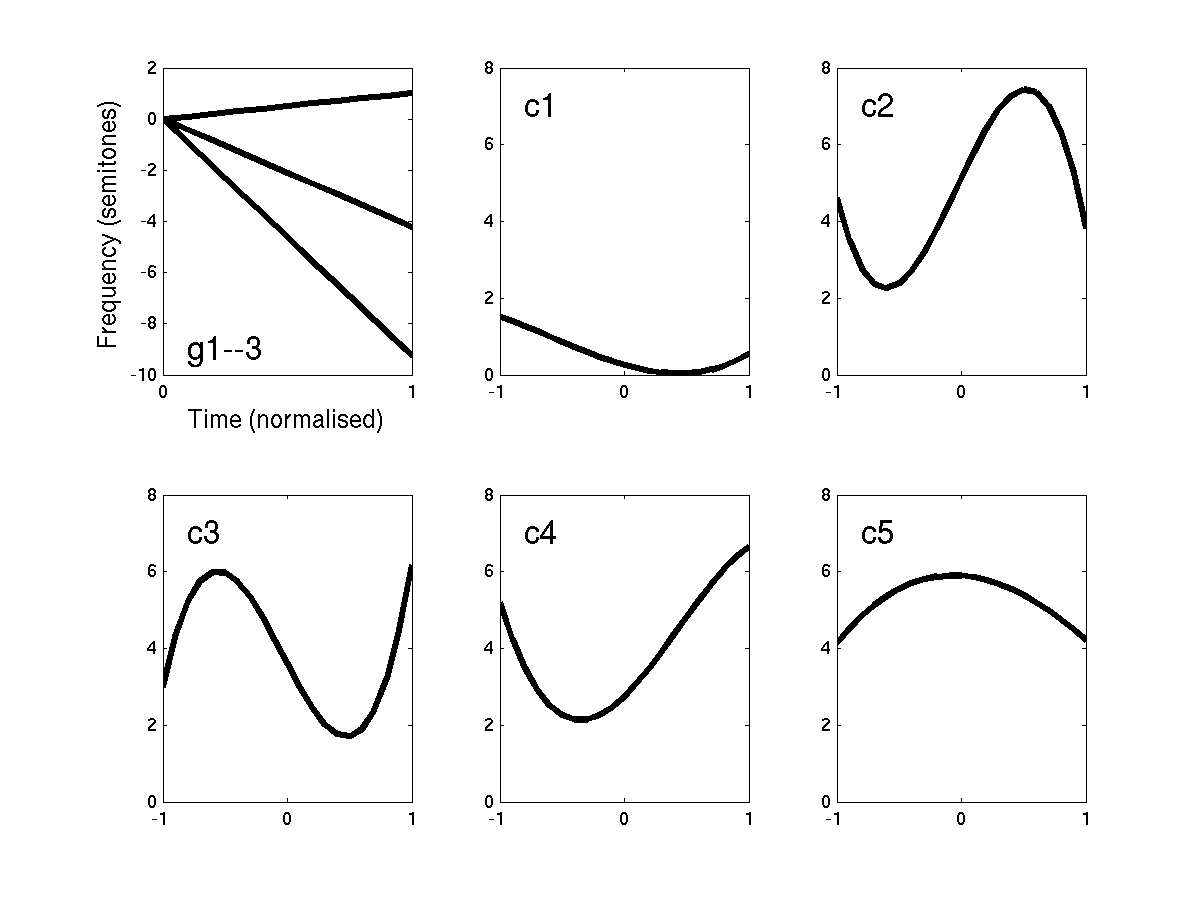}
  \caption{Global and local intonation contour classes derived by clustering.}
  \label{fig:class}
\end{figure}

Global contour classes for analyses on the categorical level are
derived by slope clustering. The cluster method can be chosen by
\myRef{clst:glob:mtd}. If the user expects a certain number of
classes, this number can be specified by
\myRef{clst:glob:kMeans:n\_cluster}.  Otherwise, meanShift clustering
should be chosen, either as the cluster method, or in combination with
kmeans for the sake of centroid initialization. For customizing the
clustering settings by non-default values several parameters are
provided whose values are passed on to the respective Python {\em
  sklearn} functions. These parameters are named as in {\em
  sklearn}. If needed, please consult the descriptions of the sklearn
functions {\em KMeans, MeanShift}, and {\em
  estimate\_bandwidth}. Figure \ref{fig:class} gives an example for
global and local contour classes.

\spar{do\_clst\_glob}{glob}{clst:glob:*}{data:myFileIdx:myChannelIdx:glob:class}

\subsection{F0 residual}
\label{sec:residual}
Dependent on \myRef{styl:register} the influence of the global component
is removed from the f0 contour in order to derive the f0 residual for
subsequent local contour stylization. If \myRef{styl:register} is set to
{\em bl, ml}, or {\em tl}, then the base, mid, or topline is
subtracted. If the parameter is set to {\em rng}, each f0 point is
normalized to the local f0 range: the corresponding points on the
base- and topline are set to 0 and 1, respectively. Thus f0 values
between base- and topline are within the range $[0$ $1]$, f0 values
below the baseline are $<0$, and values above the topline are
$>1$. For \myRef{styl:register=none} no global component influence is
removed.

\subsection{Local segments}
\label{sec:loc}

\subsubsection{Annotation}
\label{sec:locseg_annot}

In the annotation files local segments can be defined in 3 ways:
\begin{enumerate}
\item by start and end (segment tier input specified in
  \myRef{fsys:loc:tier\_ag})
\item by a center (event tier input specified in
  \myRef{fsys:loc:tier\_acc})
\item by both (segment + event tier input)
\end{enumerate}

For case (2) time stamps are transformed to segments by placing a
symmetric window of length \myRef{preproc:point\_win} on each time
stamp. In order to be able to compare contours across different
segment lengths, for (1) and (2) time is normalized as specified in
\myRef{styl:loc:nrm}. \myRef{styl:loc:nrm:mtd=minmax} yields a minmax time
normalization to the range \myRef{styl:loc:nrm:rng}.

For (3) the time stamp within the segment is treated as the
zero-center, that is, time is $[-1$ $0[$ normalized from the segment
start to the center, and $[0$ $1]$ normalized from the center to
the segment end.

For (3) only those segments in tier \myRef{fsys:loc:tier\_ag} are
considered for feature extraction to which at least one center is
assigned in tier
\myRef{fsys:loc:tier\_acc}. \myRef{preproc:loc\_align} serves for a
robus treatment of multiple center assignments. Setting this option to
{\em skip} segments with more than one center are skipped. By {\em
  left} the first center is kept, by {\em right} the last one.

\subsubsection{Contour stylization}
The f0 residual contour (see section \ref{sec:residual}) in each local
segment is stylized by n-th order polynomials.  The order is given by
\myRef{styl:loc:ord}.

\begin{figure}[!ht]
  \centering
  \includegraphics[width=8cm]{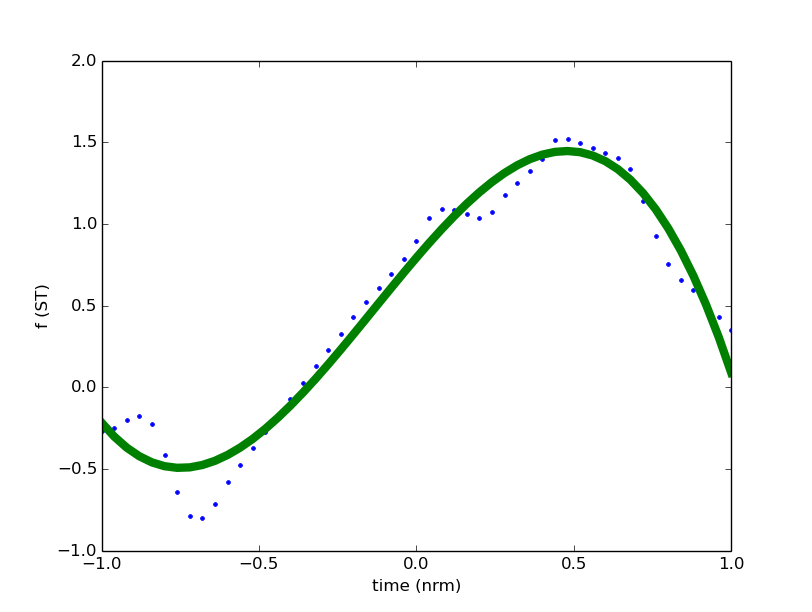}
  \caption{Local contour stylization by means of a 3rd order polynomial. Time is normalized to the range $[0$ $1]$. }
  \label{fig:loc}
\end{figure}

\begin{figure}[!ht]
  \centering
  \includegraphics[width=8cm]{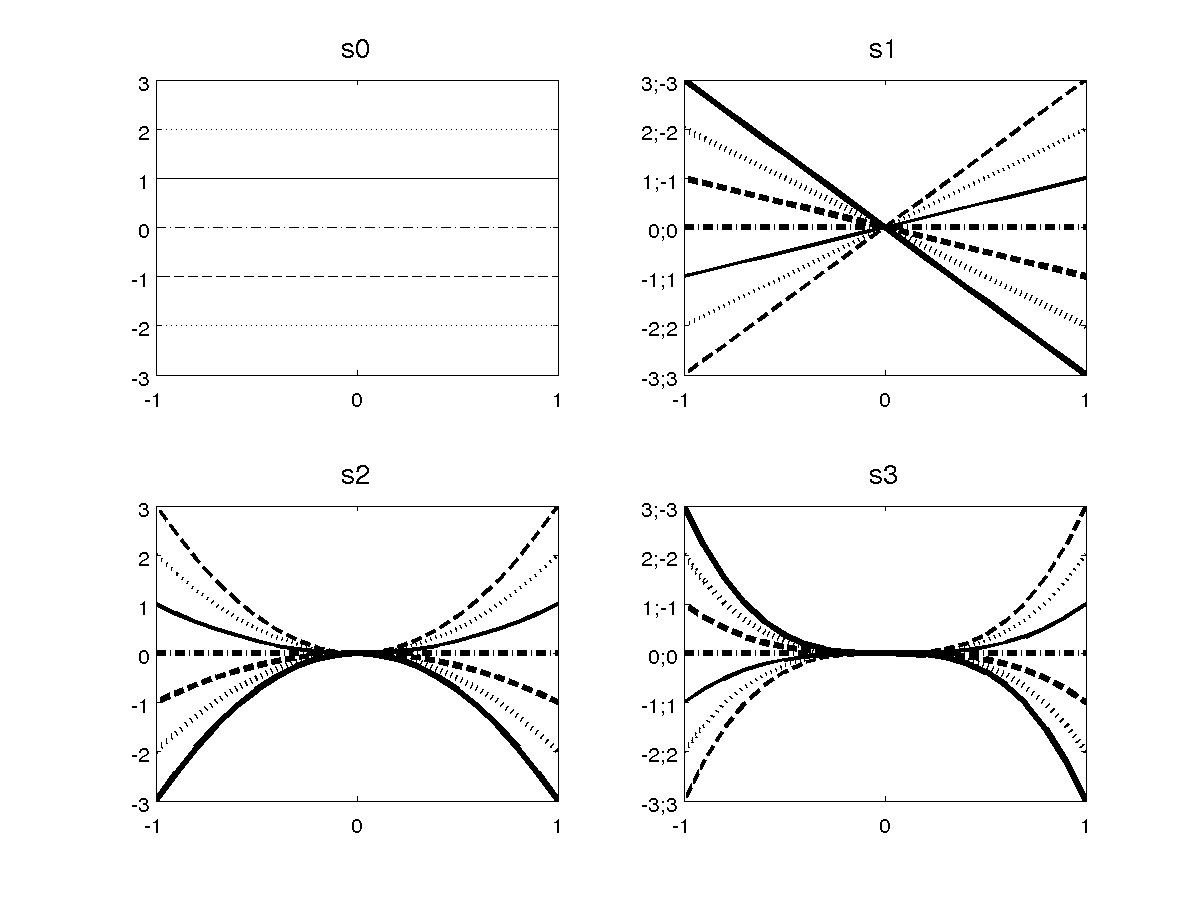}
  \caption{Influence of each coefficient of the third order polynomial
    $\sum_{i=0}^3 s_i \cdot t^i$ on the local contour shape. All other
    coefficients set to 0. For compactness purpose on the y-axis both
    function and coefficient values are shown if they differ.}
  \label{fig:polycoef}
\end{figure}

As can be seen in Figure \ref{fig:polycoef} the polynomial
coefficients are related to several aspects of local f0 shapes. Given
the polynomial $\sum_{i=0}^3 s_i \cdot t^i$, $s_0$ is related to the
local f0 level relative to the register level. $s_1$ and $s_3$ are
related to the local f0 trend (rising or falling) and -- for
annotation cases (2) and (3) -- to peak alignment, that is negative
values indicate early, and positive values late peaks. $s_2$
determines the peak shape (convex or concave) and its acuity: positive
$s_2$ values indicate convex (falling-rising) shapes, negative values
concave (rising-falling) shapes, and high values indicate stronger
acuity.

\spar{do\_styl\_loc}{loc}{styl:loc:*; styl:register}{data:myFileIdx:myChannelIdx:loc:acc:*}

\subsubsection{Contour classes}
\label{sec:clst_loc}

Local contour classes for analyses on the categorical level are
derived by polynomial coefficient clustering. The cluster method can
be chosen by \myRef{clst:loc:mtd}. If the user expects a certain number
of classes, this number can be specified by {\em
  clst:loc:kMeans:n\_cluster}.  Otherwise, meanShift clustering should
be chosen, either as cluster method, or in combination with kmeans for
the sake of centroid initialization. For customizing the clustering
settings by non-default values several parameters are provided whose
values are passed on to the respective Python {\em sklearn}
functions. These parameters are named as in {\em sklearn}. If needed,
please consult the descriptions of the {\em sklearn} functions {\em
  KMeans, MeanShift}, and {\em estimate\_bandwidth}. Figure
\ref{fig:class} gives an example for global and local contour classes.

\spar{do\_clst\_loc}{loc}{clst:loc:*}{data:myFileIdx:myChannelIdx:loc:class}

\subsubsection{Standard features}

Standard f0 and energy features are e.g. mean, standard deviation,
median, interquartile range, and maximum. They will be calculated for
the f0 contours for local contour segments. Additionally, the feature
values are locally normalized within a window of length
\myRef{preproc:nrm\_win}. See section \ref{sec:win} for window length
specifications in dependence of the annotation tier type.

\spar{do\_styl\_loc\_ext}{loc}{styl:gnl\_*:*}{data:myFileIdx:myChannelIdx:loc:gnl:*}

\subsubsection{Register features}

As with global segments, register features can also be extracted for
local features exactly the same way as introduced in section
\ref{ssec:register}.

\spar{do\_styl\_loc\_ext}{loc}{styl:glob:*}{data:myFileIdx:myChannelIdx:loc:decl:*}

\subsubsection{Gestalt features}

\begin{figure}[!ht]
  \centering
  \includegraphics[width=8cm]{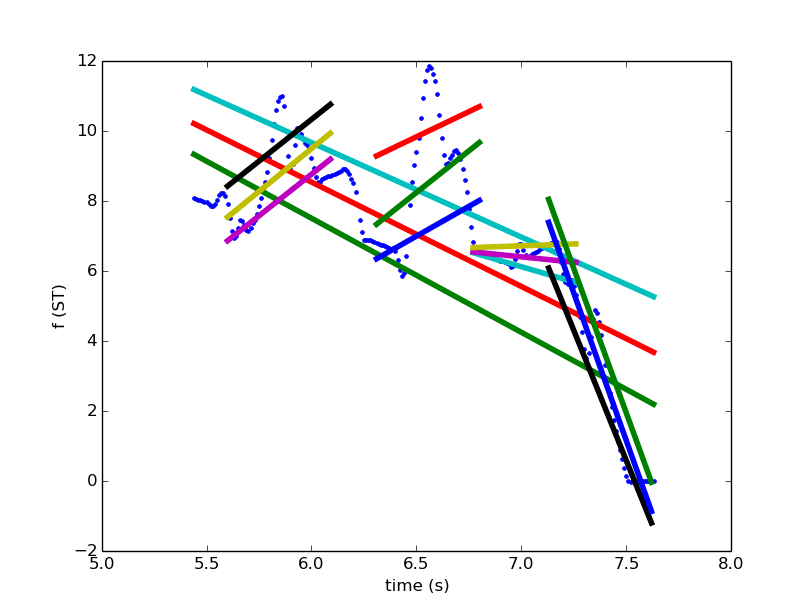}
  \caption{Gestalt stylization: Deviation of the local contour
    register aspects (base, mid, topline, range) from the global
    contour register.}
  \label{fig:gst}
\end{figure}

Gestalt features quantify the deviation of the local contour register
from the global contour register as shown in Figure \ref{fig:gst}. For
this purpose the register properties of the local segment are compared
with the properties of the dominating global segment in terms of root
mean squared deviations and slope differences. For each register
representation (base-, mid-, topline, and range regression line), the
RMSD between the local and global declination line is calculated. The
higher these values, the more the local contour sticks out from the
global contour, which is of relevance for studies on prominence,
accent group patterns \cite{BenusOLINCO2014}, and prosodic headedness
\cite{RMBIs2015, RMB_sp2018}.

The inherent Gestalt properties of the local contours are represented
again in terms of polynomial coefficients. For this purpose
polynomials of n-th order specified by \myRef{styl:loc:ord} are fitted
to all supported kinds of f0 residuals: subtraction of base-, mid-,
and topline, and range normalization. This yields 4 coefficient
vectors, one for each residual.

\spar{do\_styl\_loc\_ext}{loc}{styl:loc:*}{data:myFileIdx:myChannelIdx:loc:gst:*}

\subsection{Standard features}

Standard features are e.g. mean, standard deviation, median,
interquartile range, and maximum. They will be calculated for f0 and
energy contours over the entire file and for segments in an arbitrary
number of annotation tiers specified in \myRef{fsys:gnl\_f0:tier} and
\myRef{fsys:gnl\_en:tier}, respectively. For event tiers, the segments
are given by centering an analysis window of length
\myRef{preproc:point\_win} on the time stamps. Additionally, the
feature values are locally normalized within a window of length
\myRef{preproc:nrm\_win}. See section \ref{sec:win} for window length
specifications in dependence of the annotation tier type.
Furthermore, f0 and energy quotients are calculated between the mean
values derived in contour initial and final windows and in the
respective remainder part of the contour.  The length of this window
is specified by \myRef{styl:gnl:win}. Finally, a second order polynomial
is fitted through the f0 or energy contour, for which time is
normalized to the range $[0$ $1]$.

\subsubsection{For f0 contours}
\spar{do\_styl\_gnl\_f0}{gnl\_f0, gnl\_f0\_file}{styl:gnl\_f0:*}{data:myFileIdx:myChannelIdx:gnl\_f0:*, data:myFileIdx:myChannelIdx:gnl\_f0\_file:*}

\subsubsection{For energy contours}

An additional standard feature for energy only is spectral balance. It
is realized as the SPLH--SPL measure, i.e. the signal's sound pressure
level subtracted from the level after pre-emphasis. Pre-emphasis can
be carried out in the time of frequency domain
\myRef{styl:gnl\_en:sb:domain}. The latter is implemented as proposed
by \cite{Fant2000}. In the time domain pre-emphasis is calculated as
follows: $s'[i] = s[i]-\alpha \cdot s[i-1]$. $\alpha$ is set by
\myRef{styl:gnl\_en:sb:alpha} and determines the lower frequency
boundary for pre-emphasis by 6dB per octave. 0.95 roughly corresponds
to 150 Hz; the smaller the value for $\alpha$, the higher the lower
boundary.  Alternatively, $\alpha$ can be set directly to the lower
frequency boundary $F$ and will be internally transformed to $\alpha =
e^{-2\cdot\pi\cdot F\cdot\Delta t}$. Note that pre-emphasis in the
time domain usually leads to an overall lower energy so that SPLH--SPL
will be negative.

In the frequency domain pre-emphasis is carried out according to
\cite{Fant2000} by adding $10 \cdot
\log_{10}((1+\frac{f^2}{200^2})/(1+\frac{f^2}{5000^2}))$ to the
logarithmic spectrum.

The spectral balance calculation can be restricted to a specified time
and/or frequency window. The time window length is specified by
\myRef{styl:gnl\_en:sb:win} to cut out the center of that length of
the segment to be analysed. It serves to reduce the influence of
coarticualtion on the results. High-, low- or band-pass cutoff
frequencies (\myRef{styl:gnl\_en:sb:f}; filter type:
\myRef{styl:gnl\_en:sb:btype}) might be used to limit the analysis to
a specified frequency-band (e.g. an upper cutoff frequency 5000 Hz for
vowels).

\spar{do\_styl\_gnl\_en}{gnl\_en, gnl\_en\_file}{styl:gnl\_en:*}{data:myFileIdx:myChannelIdx:gnl\_en:*, data:myFileIdx:myChannelIdx:gnl\_en\_file:*}

\subsection{Boundaries}
\label{sec:bnd}
Boundaries are parameterized in terms of discontinuity features of
several register representations. Details and an application for
perceived prosodic boundary strength prediction can be found in
\cite{ReichelMadyIS2014}.

Boundary features can be extracted for any number of segment or event
tiers specified by \myRef{fsys:bnd:tier}. Features can be extracted for:

\begin{enumerate}
\item \myRef{navigate:do\_styl\_bnd}: each adjacent segment pair. For
  event tiers, segments are defined as the intervals between two time
  stamps. Note that this implies, that pause length is only available
  for segment tier input, where it is defined as the gap between the
  second segment's starting point and the first segment's endpoint.
\item \myRef{navigate:do\_styl\_win}: fixed time windows. For segment
  tiers, the pre- and post-boundary units are not given by the
  adjacent segments, but by windows of fixed length each of half of
  the value of \myRef{styl:bnd:win}. For event tiers the window halfs
  of \myRef{preproc:point\_win} centered on a time stamp are
  considered as pre- and post-boundary units.
\item \myRef{navigate:do\_styl\_trend}: pre- and post-boundary units,
  that range from the current chunk start to the boundary, and from
  the boundary to the chunk end. If no chunking available, the file
  start and endpoint are taken.
\end{enumerate}

For cases (2) and (3) holds: If \myRef{styl:bnd:cross\_chunk} is set to
0, and if a chunk tier is given by \myRef{fsys:tier:chunk}, the analyses
windows are limited by the start and endpoint of the current chunk.

\begin{figure}[!ht]
  \centering
  \includegraphics[width=8cm]{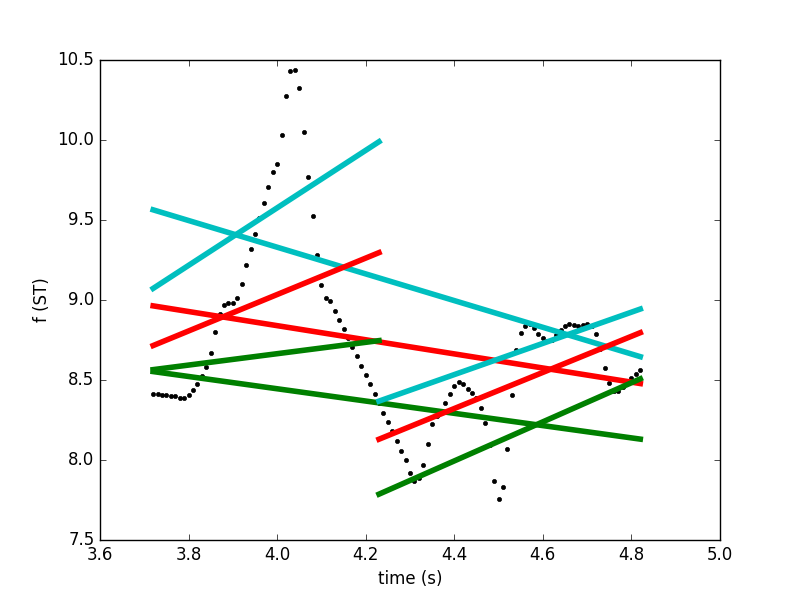}
  \caption{Prosodic boundaries: threefold base-, mid-, and topline
    register stylization for the pre-boundary, post-boundary, and the
    concatenated segment.}
  \label{fig:bnd_styl}
\end{figure}

\begin{figure}[!ht]
  \centering
  \includegraphics[width=8cm]{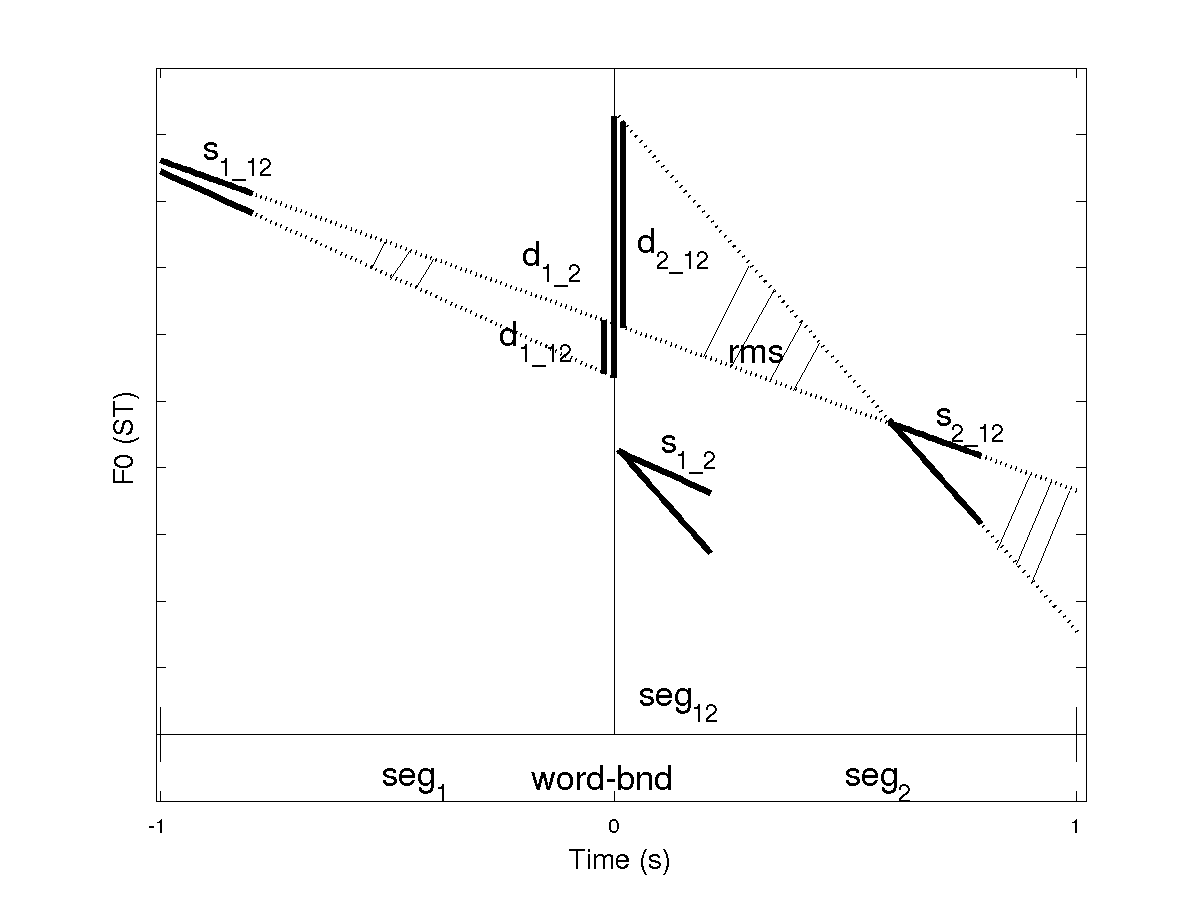}
  \caption{Boundary features describing reset and deviation from a
    common trend. In this case features are extracted at a word
    boundary {\em wrd-bnd}. The 3 regression lines can refer to f0
    baselines, midlines, toplines, and to range. The same features
    are outputted for these 4 register aspects.}
  \label{fig:bnd}
\end{figure}

A boundary is parameterized in terms of pause length (for segment tier
input only) and pitch discontinuities. For the latter, register
features (as described in section \ref{ssec:register}) are extracted
three times: for the pre-boundary segment, for the post-boundary
segment, and for the concatenation of both segments.  Figure
\ref{fig:bnd_styl} illustrates the threefold register stylization for
the pre- and post-boundary as well as for the concatenated segment.
Figure \ref{fig:bnd} shows, how discontinuity for each of the register
lines is expressed. Let seg$_1$, seg$_2$ be the pre- and post-boundary
segments, and seg$_{12}$ their concatenation. Then discontinuity is
given by:

\begin{itemize}
\item the RMSD between the four register representations of seg$_1$
  and the corresponding part of seg$_{12}$. The register
  representations are base-, mid-, topline, and range regression line.
\item the RMSD between the register representations of seg$_2$ and the
  corresponding part of seg$_{12}$
\item the RMSD between the register representations of seg$_1$ and
  seg$_2$ opposed to seg$_{12}$
\item the reset $d_{1\_2}$, i.e. the difference between the initial value of the
  regression line in seg$_2$ and the final value of the regression
  line in seg$_1$
\item the onset difference of the regression lines $d\_o$, i.e. the
  initial value of the seg$_2$ regression line subtracted from the
  initial value of the seg$_1$ line
\item the difference of the regression line mean values $d\_m$, the
  seg$_2$ mean being subtracted from the seg$_1$ mean. Both $d\_o$ and
  $d\_m$ could be used to measure downstep.
\item the pairwise slope differences $s_*$ between the 3 regression
  lines: for $s_{1\_2}$ the seg$_2$ is subtracted from the seg$_1$
  slope. For $s_{12\_1}$ and $s_{12\_2}$ the slopes of seg$_1$ and
  seg$_2$ are subtracted from the seg$_{12}$ slope.
\item the correlation-based distances between the fitted lines
  calculated for the same combinations as the RMSD values above.
  Pearson r correlations are turned into distance $d$ values ranging from
  0 to 1 by $d=\frac{1-r}{2}$.
\item the quotient of RMS errors between stylization input (the
  respective sequence of medians) and output (the fitted lines). The
  error of the joint stylization is divided by the error from the
  single pre- and post boundary fits. The quotient is reported
  separately for the entire, the pre-boundary, and the post-boundary
  segment.
\item the increase of the Akaike information criterion (AIC) resulting
  from one joint vs two separate fits. The AIC does not only account
  for the fitting error but also for the number of model
  parameters. The lower its value, the better the model. For least
  squares fit comparisons the AIC can be calculated as: $2 \cdot k + n
  \cdot \ln \textrm{RSS}$. $k$ denotes the number of model parameters,
  $n$ the number of stylization input values, and RSS the residual sum
  of squares. To each fitted line 3 parameters are assigned:
  intercept, slope, and Gaussian noise variation. The AIC increase is
  measured by subtracting the single line fit AIC from the joint fit
  AIC. It is reported separately for the entire, the pre-boundary, and
  the post-boundary segment.
\end{itemize}

All features are calculated 4 times, for the base-, mid- and toplines,
as well as for the range regression lines.

All but the reset and the slope difference variables are positively
related to discontinuity. The user might want to replace the reset and
slope differences by their absolute values.

In the \myRef{styl:bnd} option sub-dictionary {\em nrm, decl\_win}, and
{\em prct} have the same purpose right as in the \myRef{styl:glob}
context, see section \ref{ssec:register}. \myRef{styl:bnd:win} specifies
the window length of seg$_{12}$ for window case (2).

The boundary feature extraction can be carried out on the
(preprocessed) f0 contour or on the f0 residual by setting
\myRef{styl:bnd:residual} to 0 or 1, respectively. The former should
be used if boundaries between global segments as intonation phrases
are examined. The residual might be used if the user is interested in
boundaries between e.g. accent groups within the same global
segment. Note that for residuals the boundary examination across
global segments might not be meaningful, since at these boundaries the
residuals are derived from different register regression lines. These
cases can be identified in the output by means of the {\em is\_fin}
column (see section \ref{sec:otf}). The residual calculation is
described in section \ref{sec:residual}. Running boundary stylization
on residuals requires a previous global contour stylization,
i.e. \myRef{styl:navigate:do\_styl\_glob} needs to be set to true.

The subsequent paragraphs name the configuration branches associated
to the stylization cases (1)--(3), respectively.

\subsubsection{Of adjacent segments}

\spar{do\_styl\_bnd}{bnd}{styl:bnd:*}{data:myFileIdx:myChannelIdx:bnd:std:*}

\subsubsection{For fixed-length windows}

\spar{do\_styl\_bnd\_win}{bnd}{styl:bnd:*}{data:myFileIdx:myChannelIdx:bnd:win:*}

\subsubsection{For global trends}

\spar{do\_styl\_trend}{bnd}{styl:bnd:*}{data:myFileIdx:myChannelIdx:bnd:trend:*}

\subsection{Rhythm}

Rhythm features can be extracted for any number of segment or event
tiers specified by \myRef{fsys:rhy\_*:tier}, * representing {\em f0} and
{\em en} for the f0 and the energy contour, respectively.  Time stamps
of event tiers are transformed to segments as introduced in section
\ref{sec:win}.

Rhythm measures consist of:

\begin{itemize}
\item spectral moments of a DCT analysis of the contour
\item the number of peaks in the absolute-value DCT spectrum
\item the frequency associated with the highest peak
\item event rates within the analyzed segment
\item the influence of these events on the f0 or energy contour within
  the analyzed segment
\end{itemize}

To extract the relative weight of the low- and high-frequency
components of a contour, a discrete cosine transform (DCT) is applied
on the contour as in \cite{HeinrichJASA2014}. For the absolute DCT
coefficient values the first $n$ \myRef{rhy\_*:rhy:nsm} spectral
moments are calculated that (up to the forth moment) give the mean,
variance, skew, and kurtosis of the DCT coefficient weight
distribution, repsectively.

Before applying the DCT the contour is weighted by the two parameters
\myRef{rhy\_*:rhy:wintyp} and \myRef{rhy\_*:rhy:winparam} as introduced in
section \ref{ssec:en}.

The events (time stamps or segments) for which rate and influence is
to be calculated are read from one or more tier names in
\myRef{fsys:rhy\_*:tier\_rate}. Thereby within each recording channel
each {\em analysis tier} in \myRef{fsys:rhy\_*:tier} is combined with
each {\em rate tier} in \myRef{fsys:rhy\_*:tier\_rate}. Rate is simply
measured by counting the events, that fall within the segment of
analysis, and dividing it by the length of the analyzed segment. For
segment tiers in \myRef{fsys:rhy\_*:tier\_rate} only proportions
included in the segment of analysis are added to the count.

The influence $s$ of events on the f0 or energy contour is quantified
as the relative weight of the DCT coefficients around the event rate
$r$ ($+/-$ \myRef{rhy\_*:rhy:wgt:rb} Hz) within all coefficients between
\myRef{rhy\_*:rhy:lb} and \myRef{rhy\_*:rhy:ub} Hz as follows:

\begin{eqnarray}
s & = & \frac{\sum_{c: r-1 \leq f(c) \leq r+1 \textrm{Hz}} |c|}{\sum_{c: lb \leq f(c)\leq ub \textrm{Hz}} |c|} \nonumber
\end{eqnarray}

\begin{figure}[!ht]
  \centering \includegraphics[width=8cm]{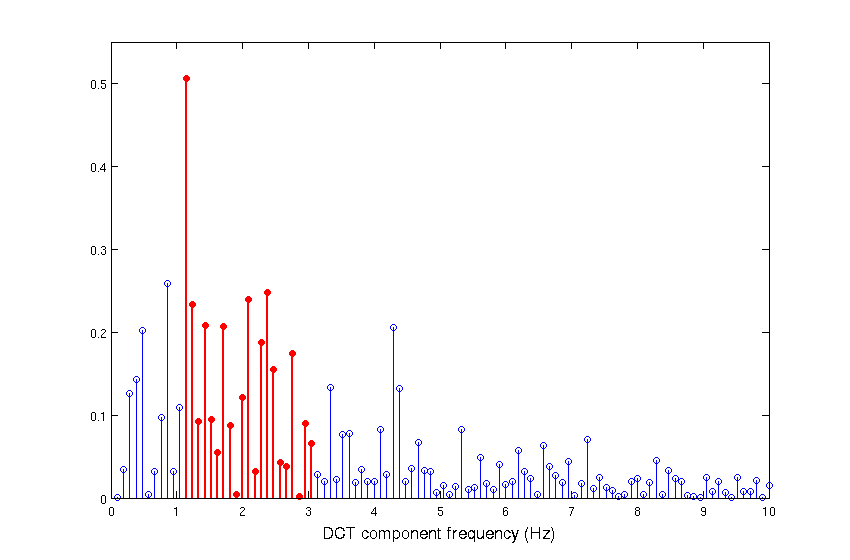}
  \hfill \includegraphics[width=8cm]{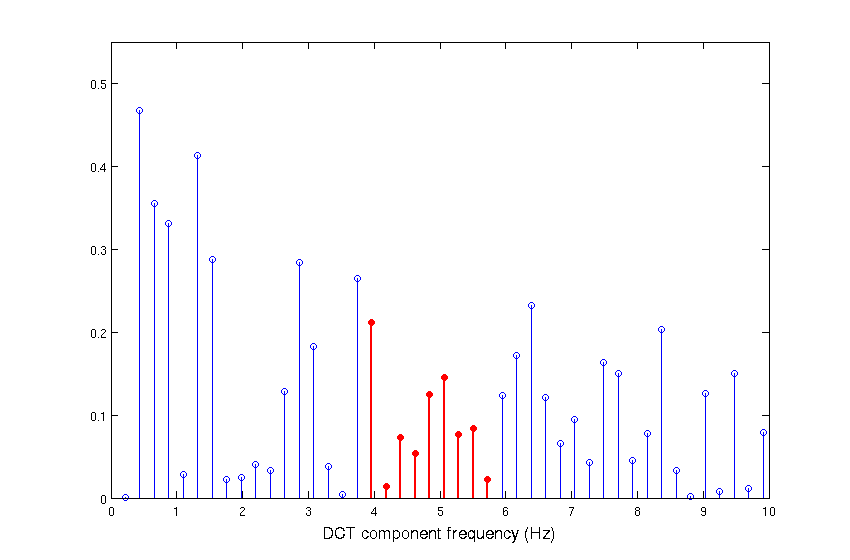}
  \caption{Influence of events on a contour in terms of the relative
    weight of the DCT coefficients around the event frequency.}
  \label{fig:rhy}
\end{figure}

The higher $s$ the higher thus the influence of the event rate on the
f0 or energy contour. Figure \ref{fig:rhy} compares a low event rate
with a high impact on the energy contour with a high event rate with
low impact (high vs low absolute coefficient values).

The relative weight is outputted to the feature table's columns {\em
  myRateTier\_prop} (see sections \ref{sec:feat} and
\ref{sec:otf}). {\em myRateTier} refers to each entry in
\myRef{fsys:rhy\_*:tier\_rate}. The respective analysis tiers from
\myRef{fsys:rhy\_*:tier} are displayed in the {\em tier} column. The
proportion is outputted for each segment in the analysis tiers.

Additionally, the rate of rate tier events in each analysis tier
segment is provided by {\em myRateTier\_rate}. Finally, {\em
  myRateTier\_mae} gives the mean absolute error between the original
contour and the inverse cosine transform output that is based on the
coefficients with frequencies around the event rates. The following
paragraphs name the configuration branches responsible for the
rhythmic analyses of the f0 and energy contour, respectively.

{\em myRateTier\_*} parameters are not calculated for analysis/rate
tier combinations across recording channels. That is: Given are
analysis tier {\em TA1} and rate tier {\em RT2} refering to channels 1
and 2, respectively. Then cells in the {\em RT2\_*} columns are set to
{\em NA} in all {\em TA1} rows, which are identified by the {\em tier}
column.

The number of peaks {\em n\_peak} in the DCT spectrum is derived by
counting the local amplitude maxima in this spectrum among the values
greater or equal than the amplitude related to the center of gravity.


\subsubsection{Rhythmic aspects of the f0 contour}

\spar{do\_styl\_rhy\_f0}{rhy\_f0, rhy\_f0\_file}{styl:rhy\_f0:*}{data:myFileIdx:myChannelIdx:rhy\_f0:*; data:myFileIdx:myChannelIdx:rhy\_f0\_file:*}

\subsubsection{Rhythmic aspects of the energy contour}

The energy contour extraction in the analyzed segment is controlled by
the \myRef{styl:rhy\_en:sig:*} sub-dictionary the same way as explained
in section \ref{ssec:en}.

\spar{do\_styl\_rhy\_en}{rhy\_en, rhy\_en\_file}{styl:rhy\_en:*}{data:myFileIdx:myChannelIdx:en\_f0:*; data:myFileIdx:myChannelIdx:rhy\_en\_file:*}

\section{Voice quality}
\label{sec:voice}

Voice quality features can be extracted for any number of segment or
event tiers specified by \myRef{fsys:voice:tier}.  Time stamps of
event tiers are transformed to segments as introduced in section
\ref{sec:win}. At the current state voice measures consist of:

\begin{itemize}
\item jitter,
\item shimmer,
\item 3rd order polynomial coefficients describing the changes of
  jitter over time
\item 3rd order polynomial coefficients describing the changes of
  shimmer over time
\end{itemize}

Note that these values are meaningful for certain domains only,
e.g. for vowel segments!

Jitter is calculated the same way as Praat's {\em relative local
  jitter}%
\footnote{\url{http://www.fon.hum.uva.nl/praat/manual/PointProcess__Get_jitter__local____.html}}
as the mean absolute difference between adjacent periods
divided by the overall mean period. As for Praat the following
parameters can be specified in \myRef{styl:voice:jit}. {\em t\_min}
and {\em t\_max} refer to the minimum and maximum allowed period
durations, and {\em fac\_max} to the maximally allowed quotient of
adjacent periods. Periods not fulfilling these constraints are
discarded from calculation.

Shimmer again is calculated the same way as Praat does for the {\em
  Shimmer (local)} parameter,%
\footnote{\url{http://www.fon.hum.uva.nl/praat/manual/Voice_3__Shimmer.html}}
i.e. it is the mean absolute difference
between the amplitudes of adjacent periods, divided by the average
amplitude.

For both jitter and shimmer a 3rd order polynomial is fitted through
the obtained sequence of distance values of adjacent periods each
distance divided by the average period, resp. amplitude. Time is
normalized to the interval -1 to 1. The purpose of these polynomials
is to represent the changes of jitter and shimmer over time. As an
example a negative 1st order coefficient for the jitter sequence
indicates a decrease in jitter over time (see Figure
\ref{fig:polycoef} for the interpretation of the coefficients).

The configuration branches related to the {\em voice} feature set are:

\spar{do\_styl\_voice}{voice, voice\_file}{styl:voice:*}{data:myFileIdx:myChannelIdx:voice:*; data:myFileIdx:myChannelIdx:voice\_file:*}

\section{Feature sets}
\label{sec:feat}

All features are subdivided into the following sets which can be extracted
independently of each other. In the subsequent listing \myRef{*\_file}
indicates that there is an additional feature extraction on the entire
file level with minor deviations from the extraction on smaller
domains (e.g. missing normalization).

\begin{itemize}
\item {\bf gnl\_f0, gnl\_f0\_file:} general standard f0 features as
  mean, median, standard deviation, interquartile range; for any number
  of tiers
\item {\bf gnl\_en, gnl\_en\_file:} general standard energy features
  as mean, median, standard deviation, interquartile range; for any
  number of tiers
\item {\bf glob:} register (level and range) features in larger
  domains (e.g. intonation phrases); for one tier per channel
\item {\bf loc:} shape features in smaller domains (e.g. accent
  groups) of f0 residuals (after removal of global f0
  aspects). Gestalt features, i.e. deviation of accent groups from
  intonation phrases. This feature set requires the precedent
  extraction of the {\em glob} set; for one tier per channel
\item {\bf bnd, bnd\_win, bnd\_trend:} boundary features between
  adjacent segments in the same domain. For {\em bnd} the features are
  derived from the stylization of adjacent segments. In {\em bnd\_win}
  the stylization is carried out in uniform time windows centered on
  the segment boundaries irrespective of the segment lengths. In {\em
    bnd\_trend} the stylization is carried out from the beginning of a
  speech chunk to the boundary in question, and from this boundary to
  the end of the chunk; for any number of tiers
\item {\bf rhy\_f0, rhy\_f0\_file:} DCT-based rhythm features; rates
  of prosodic events (e.g. syllable nuclei, pitch accents) and their
  influence on the f0 contour; for any number of tiers
\item {\bf rhy\_en, rhy\_en\_file:} DCT-based rhythm features; rates
  of prosodic events and their influence on the energy contour; for
  any number of tiers
\item {\bf voice, voice\_file:} voice quality features as jitter and
  shimmer: mean values and polynomial stylization of their changing
  over time
\end{itemize}

\paragraph{Application examples} for these feature sets are

\begin{center}
  \begin{tabular}{l|l}
    application & feature sets \\
    \hline
    pitch accent prototypes for information status and discourse segmentation \cite{ReichelCSL2014} & glob, loc \\
    prosodic boundary strength prediction \cite{ReichelMadyIS2014} & bnd \\
    prosodic typology \cite{RMBIs2015, RMB_sp2018} & loc \\
    empirical evidence for prosodic constituents (accentual phrases) \cite{BenusOLINCO2014, RMB_sp2018} & loc \\
    interplay of phrasing and prominence \cite{RMK_ESSV2016} & loc, bnd, gnl\_en, gnl\_f0 \\
    dialog act prediction \cite{MR_ESSV2016} & glob, loc, gnl\_f0 \\
    personality trait prediction \cite{ReichelIcphs2015} & glob, loc, gnl\_f0 \\
    infant-directed speech \cite{MRSKD_sp2018, mady2020primary} & glob, loc, gnl\_f0, gnl\_en \\ 
    entrainment \cite{RC_PP16, rbm_sc18} & glob, loc \\
    cooperative vs competitive dialogs \cite{rl_essv2018} & glob, loc, rhy\_en \\
    offtalk detection \cite{MR_PP16} & glob, loc, gnl\_en, gnl\_f0 \\
    speech disfluencies \cite{BelzReichelDiss2015} & loc \\
    pitch accent inventory for low-resource languages \cite{Kalkhoff2015} & loc \\
    Lombard speech characteristics \cite{BRS_IS2015} & bnd \\
    Social media analyses \cite{RL_WNUT2016} & bnd \\
    Hand-stroke--speech coordination \cite{FR_PP16} & rhy\_en, rhy\_f0 \\
    Acoustics of non-lexical speech \cite{rkm2023} & glob, loc, gnl\_en, gnl\_f0
  \end{tabular}
\end{center}

The following tables list all currently available features in
alphabetical order, give short descriptions and link them to the
respective feature set.  In these tables {\em loc} and {\em glob}
within the superpositional setting refer to local (e.g. accent groups)
and global segments (e.g. intonation phrases), respectively. For
boundary parameterization {\em pre, post, joint} refer to the pre- and
post-boundary segments, and to their concatenation, respectively. For
boundary features {\em std, win}, and {\em trend} refer to the
underlying windowing of neighboring segments, cf. section
\ref{sec:bnd}. The number of coefficient and spectral moment variables
    {\em c*} and {\em sm*} depend on the polynomial order and spectral
    moment number specified by the user. For the {\em rhy\_*} feature
    sets {\em myAnalysisTier} stands for the analysis tier, and {\em
      myRateTier} for the rate tier, i.e. the rate and influence of
    events in {\em myRateTier} within segments of {\em myAnalysisTier}
    is measured, and all possible combinations of analysis and rate
    tiers are outputted.


\newpage
    
\begin{small}

  \begin{tabular}{l|l|l}
    {\bf name} & {\bf description} & {\bf feature set} \\
    \hline
    bl\_c0 & baseline intercept & glob, loc \\
    bl\_c1 & baseline slope & glob, loc \\
    bl\_d & mean baseline deviation loc-glob & loc \\
    bl\_d\_fin & final baseline value diff loc-glob & loc \\
    bl\_d\_init & initial baseline value diff loc-glob & loc \\
    bl\_drop & baseline f0 drop (duration $\cdot$ rate) & glob \\
    bl\_m & baseline mean value & glob, loc \\
    bl\_r & baseline reset & glob \\
    bl\_rate & baseline declination rate & glob, loc \\
    bl\_rms & baseline RMSD loc-glob & loc \\
    bl\_sd & baseline slope diff loc-glob & loc \\
    bv & file-domain f0 base value (Hz) & glob, gnl\_f0\_file \\
    c* & polynomial loc contour coef * & loc, gnl\_f0/en(\_file) \\
    ci & channel index (starting with 0) & {\em (all sets)} \\
    class & contour class & glob, loc \\ 
    dur & segment duration & glob, loc, gnl\_f0/en(\_file), rhy\_f0/en(\_file) \\
    dur\_nrm & normalized duration & loc, gnl\_f0/en \\
    r\_en\_f0 & correlation between energy and f0 contour & gnl\_en \\
    f\_max & freq of coef with max ampl. in DCT spectrum & rhy\_f0/en(\_file) \\
    fi & file index (starting with 0) & {\em (all sets)} \\
    gi & si value of corresponding row in glob & loc \\
    iqr & f0 interquartile range & glob, loc, gnl\_f0/en(\_file) \\
    iqr\_nrm & nrm'd f0 interquartile range & loc, gnl\_f0/en \\
    is\_fin & item in global segment's final position? & {\em (all sets w/o *\_file)} \\
    is\_fin\_chunk & item in chunk final position? & {\em (all sets w/o *\_file)} \\
    is\_init & item in global segment's initial position? & {\em (all sets w/o *\_file)} \\
    is\_init\_chunk & item in chunk initial position? & {\em (all sets w/o *\_file)} \\
    jit & jitter & voice(\_file) \\
    jit\_c* & polynomial coefs for jitter time course & voice(\_file) \\
    jit\_m & mean pulse period & voice(\_file) \\
    jit\_m\_nrm & normalized mean pulse period & voice \\
    jit\_nrm & normalized jitter & voice \\
    jit\_sd & pulse period std & voice(\_file) \\
    jit\_sd\_nrm & normalized pulse period std & voice(\_file) \\
    shim & shimmer & voice(\_file) \\
    shim\_c* & polynomial coefs for shimmer time course & voice(\_file) \\
    shim\_m & mean pulse amplitude & voice(\_file) \\
    shim\_m\_nrm & normalized mean pulse amplitude & voice \\
    shim\_nrm & normalized shimmer & voice \\
    shim\_sd & pulse amplitude std & voice(\_file) \\
    shim\_sd\_nrm & normalized pulse amplitude std & voice(\_file) \\
    lab & label & glob, bnd, gnl\_f0/en, rhy\_f0/en \\
    lab\_acc & ACC tier label & loc \\
    lab\_ag & AG tier label & loc \\
    lab\_next & next segment's label & bnd \\
    m & f0, energy arit. mean & glob, loc, gnl\_f0/en(\_file) \\
    m\_nrm & f0, energy arit. nrm'd mean & loc, gnl\_f0/en \\
    max & f0, energy max & glob, loc, gnl\_f0/en(\_file) \\
    max\_nrm & f0, energy nrm'd max & loc, gnl\_f0/en \\
    maxpos & relative position of maximum & glob, loc, gnl\_f0/en \\
    med & f0, energy median & glob, loc, gnl\_f0/en(\_file) \\
    med\_nrm & f0, energy nrm'd median & loc, gnl\_f0/en \\
    ml\_bl\_cross\_f0 & f0 of crossing point of mid- and baseline & glob \\
    ml\_bl\_cross\_t & time of crossing point of mid- and baseline & glob \\
    ml\_c0 & midline intercept & glob, loc \\
    ml\_c1 & midline slope & glob, loc \\
    ml\_d & mean midline deviation loc-glob & loc \\
    ml\_d\_fin & final midline value diff loc-glob & loc \\
    ml\_d\_init & initial midline value diff loc-glob & loc \\
    ml\_drop & midline f0 drop (duration $\cdot$ rate) & glob \\
    ml\_m & midline mean value & glob, loc \\
    ml\_r & midline reset & glob \\
    ml\_rate & midline declination rate & glob, loc \\
    ml\_rms & midlines RMSD loc-glob & loc \\
    ml\_sd & midline slope diff loc-glob & loc \\
    n\_peak & number of peaks in absoulte DCT spectrum & rhy\_f0/en(\_file)
  \end{tabular}
  
\end{small}

\newpage

\begin{small}
  \begin{tabular}{l|l|l}
    p & pause length (sec) & bnd \\
    qb & quotient of means of init and fin part & gnl\_f0/en(\_file) \\
    qf & quotient of means of final and non-fin part & gnl\_f0/en(\_file) \\
    qi & quotient of means of initial and non-init & gnl\_f0/en(\_file) \\
    qm & quotient of means max(init, fin) part and remainder & gnl\_f0/en(\_file) \\
    res\_bl\_c* & baseline residual poly coef * & loc \\
    res\_ml\_c* & midline residual poly coef * & loc \\
    res\_rng\_c* & range line residual poly coef * & loc \\
    res\_tl\_c* & topline residual poly coef * & loc \\
    rms & overall RMSD & gnl\_en \\
    rms\_nrm & nrm'd overall RMSD & gnl\_en \\
    rmsd & RMSD under stylized contour & loc \\ 
    rng\_c0 & range line intercept & glob, loc \\
    rng\_c1 & range line slope & glob, loc \\
    rng\_d & mean range line deviation loc-glob & loc \\
    rng\_d\_fin & final range line value diff loc-glob & loc \\
    rng\_d\_init & initial range line value diff loc-glob & loc \\
    rng\_drop & range line f0 drop (duration $\cdot$ rate) & glob \\
    rng\_m & range mean value & glob, loc \\
    rng\_r & range line reset & glob\\
    rng\_rate & range declination rate & glob, loc \\
    rng\_rms & range lines RMSD loc-glob & loc \\
    rng\_sd & range line slope diff loc-glob & loc \\
    sb & spectral balance & gnl\_en \\
    sd & f0, energy standard deviation & glob, loc, gnl\_f0/en(\_file) \\
    sd\_nrm & nrm'd f0, energy standard deviation & loc, gnl\_f0, gnl\_en \\
    si & segment index (starting with 0) & glob, loc, gnl\_f0/en, rhy\_f0/en \\
    sm* & *th spectral moment of DCT & rhy\_f0/en(\_file)  \\
    std$|$trend$|$win\_bl\_aicI & baseline fitting AIC increase joint vs pre+post & bnd \\
    std$|$trend$|$win\_bl\_aicI\_post & baseline fitting AIC increase joint vs post & bnd \\
    std$|$trend$|$win\_bl\_aicI\_pre & baseline fitting AIC increase joint vs pre & bnd \\
    std$|$trend$|$win\_bl\_corrD & pre/post-joint baseline corr-based distance & bnd \\
    std$|$trend$|$win\_bl\_corrD\_post & post-joint baseline corr-based distance & bnd \\
    std$|$trend$|$win\_bl\_corrD\_pre & pre-joint baseline corr-based distance & bnd \\
    std$|$trend$|$win\_bl\_d\_m & difference of baseline means pre--post & bnd \\
    std$|$trend$|$win\_bl\_d\_o & difference of baseline onsets pre--post & bnd \\
    std$|$trend$|$win\_bl\_r & pre-post baseline reset & bnd \\
    std$|$trend$|$win\_bl\_rms & pre/post-joint baseline RMSD & bnd \\
    std$|$trend$|$win\_bl\_rms\_post & post-joint baseline RMSD & bnd \\
    std$|$trend$|$win\_bl\_rms\_pre & pre-joint baseline RMSD & bnd \\
    std$|$trend$|$win\_bl\_rmsR & baseline fitting error ratio joint vs pre+post & bnd \\
    std$|$trend$|$win\_bl\_rmsR\_post & baseline fitting error ratio joint vs post & bnd \\
    std$|$trend$|$win\_bl\_rmsR\_pre & baseline fitting error ratio joint vs pre & bnd \\
    std$|$trend$|$win\_bl\_sd\_post & baseline slope diff post--joint & bnd \\
    std$|$trend$|$win\_bl\_sd\_pre & baseline slope diff pre--joint & bnd \\
    std$|$trend$|$win\_bl\_sd\_prepost & baseline slope diff pre--post & bnd \\    
    std$|$trend$|$win\_ml\_aicI & midline fitting AIC increase joint vs pre+post & bnd \\
    std$|$trend$|$win\_ml\_aicI\_post & midline fitting AIC increase joint vs post & bnd \\
    std$|$trend$|$win\_ml\_aicI\_pre & midline fitting AIC increase joint vs pre & bnd \\
    std$|$trend$|$win\_ml\_corrD & pre/post-joint midline corr-based distance & bnd \\
    std$|$trend$|$win\_ml\_corrD\_post & post-joint midline corr-based distance & bnd \\
    std$|$trend$|$win\_ml\_corrD\_pre & pre-joint midline corr-based distance & bnd \\
    std$|$trend$|$win\_ml\_d\_m & difference of midline means pre--post & bnd \\
    std$|$trend$|$win\_ml\_d\_o & difference of midline onsets pre--post & bnd \\
    std$|$trend$|$win\_ml\_r & pre--post midline reset & bnd \\
    std$|$trend$|$win\_ml\_rms & pre/post--joint midline RMSD & bnd \\
    std$|$trend$|$win\_ml\_rms\_post & post-joint midline RMSD & bnd \\
    std$|$trend$|$win\_ml\_rms\_pre & pre-joint midline RMSD & bnd \\
    std$|$trend$|$win\_ml\_rmsR & midline fitting error ratio joint vs pre+post & bnd \\
    std$|$trend$|$win\_ml\_rmsR\_post & midline fitting error ratio joint vs post & bnd \\
    std$|$trend$|$win\_ml\_rmsR\_pre & midline fitting error ratio joint vs pre & bnd \\
    std$|$trend$|$win\_ml\_sd\_post & midline slope diff post--joint & bnd \\
    std$|$trend$|$win\_ml\_sd\_pre & midline slope diff pre--joint & bnd \\
    std$|$trend$|$win\_ml\_sd\_prepost & midline slope diff pre-post & bnd \\
    std$|$trend$|$win\_rng\_aicI & range fitting AIC increase joint vs pre+post & bnd \\
    std$|$trend$|$win\_rng\_aicI\_post & range fitting AIC increase joint vs post & bnd \\
    std$|$trend$|$win\_rng\_aicI\_pre & range fitting AIC increase joint vs pre & bnd \\
    std$|$trend$|$win\_rng\_corrD & pre/post-joint range line corr-based distance & bnd \\
    std$|$trend$|$win\_rng\_corrD\_post & post-joint range line corr-based distance & bnd \\
    std$|$trend$|$win\_rng\_corrD\_pre & pre-joint range line corr-based distance & bnd
  \end{tabular}
\end{small}

\newpage

\begin{small}
  \begin{tabular}{l|l|l}
    std$|$trend$|$win\_rng\_d\_m & difference of range line means pre--post & bnd \\
    std$|$trend$|$win\_rng\_d\_o & difference of range line onsets pre--post & bnd \\
    std$|$trend$|$win\_rng\_r & pre-post range line reset & bnd \\
    std$|$trend$|$win\_rng\_rms & std pre/post-joint range line RMSD & bnd \\
    std$|$trend$|$win\_rng\_rms\_post & post-joint range line RMSD & bnd \\
    std$|$trend$|$win\_rng\_rms\_pre & pre-joint range line RMSD & bnd \\
    std$|$trend$|$win\_rng\_rmsR & range fitting error ratio joint vs pre+post & bnd \\
    std$|$trend$|$win\_rng\_rmsR\_post & range fitting error ratio joint vs post & bnd \\
    std$|$trend$|$win\_rng\_rmsR\_pre & range fitting error ratio joint vs pre & bnd \\
    std$|$trend$|$win\_rng\_sd\_post & range line slope diff post-joint & bnd \\
    std$|$trend$|$win\_rng\_sd\_pre & range line slope diff pre-joint & bnd \\
    std$|$trend$|$win\_rng\_sd\_prepost & range line slope diff pre-post & bnd \\
    std$|$trend$|$win\_tl\_aicI & topline fitting AIC increase joint vs pre+post & bnd \\
    std$|$trend$|$win\_tl\_aicI\_post & topline fitting AIC increase joint vs post & bnd \\
    std$|$trend$|$win\_tl\_aicI\_pre & topline fitting AIC increase joint vs pre & bnd \\
    std$|$trend$|$win\_tl\_corrD & pre/post-joint topline corr-based distance & bnd \\
    std$|$trend$|$win\_tl\_corrD\_post & post-joint topline corr-based distance & bnd \\
    std$|$trend$|$win\_tl\_corrD\_pre & pre-joint topline corr-based distance & bnd \\
    std$|$trend$|$win\_tl\_d\_m & difference of topline means pre--post & bnd \\
    std$|$trend$|$win\_tl\_d\_o & difference of topline onsets pre--post & bnd \\
    std$|$trend$|$win\_tl\_r & std pre-post topline reset & bnd \\
    std$|$trend$|$win\_tl\_rms & pre/post-joint topline RMSD & bnd \\
    std$|$trend$|$win\_tl\_rms\_post & post-joint topline RMSD & bnd \\
    std$|$trend$|$win\_tl\_rms\_pre & pre-joint topline RMSD & bnd \\
    std$|$trend$|$win\_tl\_rmsR & topline fitting error ratio joint vs pre+post & bnd \\
    std$|$trend$|$win\_tl\_rmsR\_post & topline fitting error ratio joint vs post & bnd \\
    std$|$trend$|$win\_tl\_rmsR\_pre & topline fitting error ratio joint vs pre & bnd \\
    std$|$trend$|$win\_tl\_sd\_post & topline slope diff post-joint & bnd \\
    std$|$trend$|$win\_tl\_sd\_pre & topline slope diff pre-joint & bnd \\
    std$|$trend$|$win\_tl\_sd\_prepost & topline slope diff pre-post & bnd \\
    stm & f0 file name stem & glob, loc \\
    t\_off & time offset (sec; bnd: of pre-boundary segment) & glob, loc, gnl\_f0/en, rhy\_f0/en, bnd \\
    t\_on & time onset (sec; bnd: of post-boundary segment) & glob, loc, gnl\_f0/en, rhy\_f0/en, bnd \\
    tier & tier name & bnd, glob, gnl\_f0/en, rhy\_f0/en\\
    tier\_acc & accent point tier name & loc \\
    tier\_ag & accent group segment tier name & loc \\
    tl\_bl\_cross\_f0 & f0 of crossing point of top- and baseline & glob \\
    tl\_bl\_cross\_t & time of crossing point of top- and baseline & glob \\
    tl\_ml\_cross\_f0 & f0 of crossing point of top- and midline & glob \\
    tl\_ml\_cross\_t & time of crossing point of top- and midline & glob \\
    tl\_c0 & topline intercept & glob \\
    tl\_c1 & topline slope & glob \\
    tl\_d & mean topline deviation loc-glob & loc \\
    tl\_d\_fin & final topline value diff loc-glob & loc \\
    tl\_d\_init & initial topline value diff loc-glob & loc \\
    tl\_drop & topline f0 drop (duration $\cdot$ rate) & glob \\
    tl\_m & topline mean value & glob, loc \\
    tl\_r & initial topline reset & glob \\
    tl\_rate & topline declination rate & glob, loc \\
    tl\_rms & topline RMSD loc-glob & loc \\
    tl\_sd & topline slope diff loc-glob & loc \\
        myRateTier\_dgm & difference between rate and frequency of max amplitude coef & rhy\_f0/en(\_file) \\
    myRateTier\_dlm & difference between rate and frequency of nearest peak coef & rhy\_f0/en(\_file) \\
    myRateTier\_mae & meanAbsErr(IDCT(myRateTier),contourOfFile) & rhy\_f0/en(\_file) \\
    myRateTier\_prop & influence of myRateTier on DCT coefs & rhy\_f0/en(\_file) \\
    myRateTier\_rate & event rate of myRateTier & rhy\_f0/en(\_file)
  \end{tabular}
\end{small}

\section{Configurations}
\label{sec:config}
The configuration file format is {\em JSON}. Examples can be found in
the {\em config} subfolder of the code distribution. {\em
  copasul\_default\_config.json} contains all default values. In the
{\em doc} subfolder you find the file {\em
  copasul\_commented\_config.json.txt} where all options are commented
for a quick overview. In the following detailed introduction of all
configuration parameters, the levels of the JSON dictionary are
separated by a colon.

For numeric and boolean parameters the {\em ``values, default''} field contains
the default value. For string parameters, the default value is
indicated in bold face. If a configuration field is named as {\em my*}
the name is user defined. $+$ indicates ``one or more'' configuration
branches of this kind. Example: \myRef{fsys:channel:myTiername+}
indicates, that the user needs to specify for all tiers in the
annotation files, to which audio channel they belong. Let's assume
there are two tiers {\em spk1} and {\em spk2}, the first belongs to
channel 1, the second to channel, two, then \myRef{fsys:channel:spk1=1}
and \myRef{fsys:channel:spk2=2}.

\subsection{Sample rate}
\copar{fs}{f0 sample frequency}{integer}{100}{currently only fs=100
  supported. All f0 input will be resampled to this sample rate}

\subsection{Navigation}

\subsubsection{Augmentation}

Automatic annotation steps can be carried out independently of each
other as long they don't depend on the output of preceding annotation
steps, e.g. if fallback events as syllable boundaries and nuclei are
required for phrase boundary and accent detection, or if parent
segments are defined to be the result of preceding automatic
clustering or prosodic phrasing. Figure \ref{fig:navi_aug} displays
the possible augmentation pipelines.

\tikzstyle{proc} = [rectangle, draw, fill=blue!20, node distance=1cm and 3cm, 
  text centered, rounded corners, minimum height=1em]

\begin{figure}[!hbt]
  \begin{center}
    \begin{tikzpicture}[node distance = 2cm, auto]
      \node [proc] (clst) {chunk};
      \coordinate[above of=clst] (start);
      \node [proc, below of=clst] (syl) {syl};
      \node [proc, below of=syl] (glob) {glob};
      \node [proc, below of=glob] (loc) {loc};
      \coordinate[below of=loc] (end);
      \path[-] (start) edge (clst);
      \path[-] (start) edge[bend right=90] (syl);
      \path[-] (start) edge[bend right=90] (glob);
      \path[-] (start) edge[bend right=90] (loc);
      \path[-] (clst) edge (syl);
      \path[-] (clst) edge[bend left=90] (glob);
      \path[-] (clst) edge[bend left=90] (loc);
      \path[-] (syl) edge (glob);
      \path[-] (syl) edge[bend left=90] (loc);
      \path[-] (glob) edge (loc);
      \path[-] (loc) edge (end);
    \end{tikzpicture}
    \caption{Automatic annotation {\em do\_augment\_*} workflow}
    \label{fig:navi_aug}
  \end{center}
\end{figure}
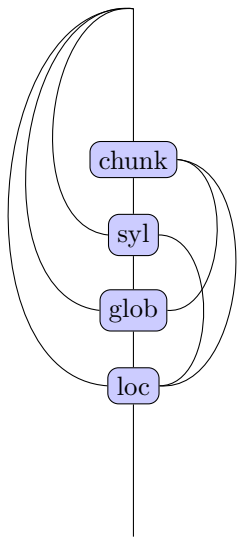

\subsubsection{Feature extraction}

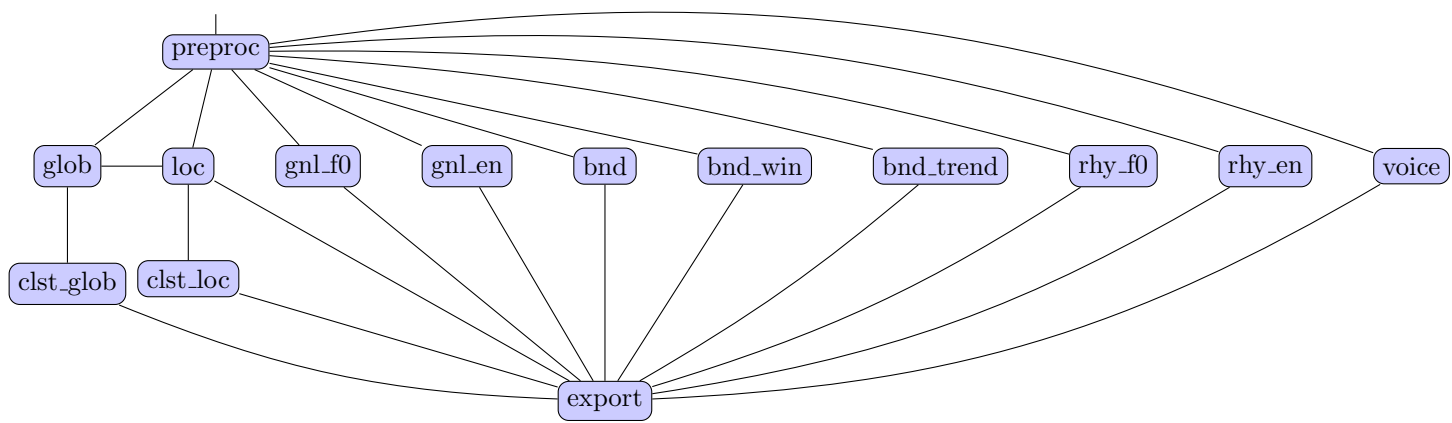
\begin{figure}[!hbt]
  \begin{center}
    \begin{tikzpicture}[node distance = 0.5cm, auto]
      \node [proc] (preproc) {preproc}; \coordinate[above of=preproc]
      (start); \node [proc, below left = 1cm and 0.8cm of preproc]
      (glob) {glob}; \node [proc, right=0.8cm of glob] (loc) {loc};
      \node [proc, below=1cm of loc] (clstLoc) {clst\_loc}; \node
            [proc, below=1cm of glob] (clstGlb) {clst\_glob}; \node
            [proc, right=0.8cm of loc] (gnlF0) {gnl\_f0}; \node [proc,
              right=0.8cm of gnlF0] (gnlEn) {gnl\_en}; \node [proc,
              right=0.8cm of gnlEn] (bnd) {bnd}; \node [proc,
              right=0.8cm of bnd] (bndWin) {bnd\_win}; \node [proc,
              right=0.8cm of bndWin] (bndTrend) {bnd\_trend}; \node
            [proc, right=0.8cm of bndTrend] (rhyF0) {rhy\_f0}; \node
            [proc, right=0.8cm of rhyF0] (rhyEn) {rhy\_en}; \node
            [proc, right=0.8cm of rhyEn] (voice) {voice}; \node [proc,
              below=2.6cm of bnd] (export) {export}; \path[-] (start)
            edge (preproc); \path[-] (preproc) edge (glob); \path[-]
            (preproc) edge (loc); \path[-] (preproc) edge (gnlF0);
            \path[-] (preproc) edge (gnlEn); \path[-] (preproc) edge
            (bnd); \path[-] (preproc) edge (bndWin); \path[-]
            (preproc) edge[bend left=5] (bndTrend); \path[-] (preproc)
            edge[bend left=8] (rhyF0); \path[-] (preproc) edge[bend
              left=11] (rhyEn); \path[-] (preproc) edge[bend left=14]
            (voice); \path[-] (glob) edge (loc); \path[-] (glob) edge
            (clstGlb); \path[-] (loc) edge (clstLoc); \path[-]
            (clstGlb) edge[bend right=10] (export); \path[-] (clstLoc)
            edge (export); \path[-] (loc) edge (export); \path[-]
            (gnlF0) edge (export); \path[-] (gnlEn) edge (export);
            \path[-] (bnd) edge (export); \path[-] (bndWin) edge
            (export); \path[-] (bndTrend) edge[bend left=5] (export);
            \path[-] (rhyF0) edge[bend left=8] (export); \path[-]
            (rhyEn) edge[bend left=11] (export); \path[-] (voice)
            edge[bend left=14] (export);
    \end{tikzpicture}

    \caption{Stylization {\em do\_styl} and clustering {\em do\_clst} workflow}
    \label{fig:navi_feat}
  \end{center}
\end{figure}

\paragraph{Processing pipelines}
Pipelines are defined in the \myRef{navigate}
configurations. Processing step dependencies are shown in Figure
\ref{fig:navi_feat}. 

Processing does not always need to start from scratch. Intermediate
feature extraction results are stored in Python pickle format and can
be reloaded for further processing in a later session. The name of the
pickle file to be loaded is given in

\begin{quote}
  \myRef{fsys:export:dir} + \myRef{fsys:export:stm}
\end{quote}

In order to continue an analysis of a previous session, the user thus
needs to make sure that output directory and file name stem do not
change across sessions. The content of the file can be deleted by
setting \myRef{navigate:from\_scratch} to true. This and all other
\myRef{navigate} configuration elements are introduced in the
following:

\copar{navigate:do\_augment\_chunk}{apply automatic chunking into interpausal units}{boolean}{false}{If true, a chunk segment tier is generated for each channel and added to the annotation files.}
\copar{navigate:do\_augment\_glob}{apply unsupervised prosodic phrase extraction}{boolean}{false}{If true, for each channel a segment tier with automatically extracted prosodic phrases is generated and added to the annotation files. If no input tier for prosodic boundary candidates is specified, this step requires preceding syllable extraction, since syllable boundaries will then be taken as candidates.}
\copar{navigate:do\_augment\_loc}{apply unsupervised pitch accent detection}{boolean}{false}{If true, for each channel an event tier with automatically extracted pitch accent locations is generated and added to the annotation file. If no user-defined pitch accent candidates can be provided, this step requires preceding syllable nucleus extraction, which will then be taken as candidates.}
\copar{navigate:do\_augment\_syl}{apply automatic syllable nucleus and boundary detection}{boolean}{false}{If true, for each channel two event tiers -- a syllable nucleus and boundary tier -- are generated and added to the annotation files.}
\copar{navigate:do\_clst\_glob}{apply local contour clustering}{boolean}{false}{cluster local contour polynomial coefficients to derive local intonation contour classes.}
\copar{navigate:do\_clst\_loc}{apply global contour clustering}{boolean}{false}{cluster global contour line slope coefficients to derive global intonation contour classes.}
\copar{navigate:do\_export}{export the results}{boolean}{false}{generate csv feature table files, and f0 table files}
\copar{navigate:do\_plot}{plot}{boolean}{false}{online or post-analysis plotting of stylization results. Online plotting serves to check the parameter settings before processing large data.}
\copar{navigate:do\_preproc}{apply preprocessing}{boolean}{false}{F0 preprocessing and analysis and normalization windowing. If set to true at non-initial application to a data set, all information previously gathered from subsequent stylization steps is deleted.}
\copar{navigate:do\_styl\_bnd\_trend}{extract boundary features}{boolean}{false}{Extract f0 discontinuity features at each segment boundary or time stamp. This time the pre- and post-boundary units range from file start to the boundary, and from the boundary to the file end. If \myRef{styl:bnd:cross\_chunk} is set to false, and if a chunk tier is given in \myRef{fsys:chunk:tier}, the analyses windows are limited by the start and endpoint of the current chunk.}
\copar{navigate:do\_styl\_bnd\_win}{extract boundary features in fixed time windows}{boolean}{false}{Extract f0 discontinuity features. For segment tiers, the pre- and post-boundary units are not given by the adjacent segments as for \myRef{navigate:do\_styl\_bnd}, but by windows of fixed length. For event tiers the window halfs of \myRef{preproc:point\_win} centered on a time stamp are considered as pre- and post-boundary units. If \myRef{styl:bnd:cross\_chunk} is set to false, and if a chunk tier is given in \myRef{fsys:chunk:tier}, the analyses windows are limited by the start and endpoint of the current chunk.}
\copar{navigate:do\_styl\_bnd}{extract boundary features}{boolean}{false}{Extract f0 discontinuity features across segments (segment tier input) or at time stamps (event tier input). Only for the former the extracted pause length is meaningful. Discontinuity is amongst others expressed in the deviation of the pre- and post-boundary part from a common declination trend. For segment tiers, this common trend is calculated over both segments. For event tiers, the inter-time stamp intervals are considered as segments.}
\copar{navigate:do\_styl\_glob}{apply global contour stylization}{boolean}{false}{Apply f0 register (level and range) stylizations within global segments as e.g. IPs.}
\copar{navigate:do\_styl\_gnl\_en}{extract standard energy features}{boolean}{false}{Extract energy mean, variance and the like.}
\copar{navigate:do\_styl\_gnl\_f0}{extract standard f0 features}{boolean}{false}{Extract f0 mean, variance and the like.}
\copar{navigate:do\_styl\_loc\_ext}{extract extended feature set for local f0 contours}{boolean}{false}{Extract local register and Gestalt features, i.e. deviation of the local contour from the global register trend.}
\copar{navigate:do\_styl\_loc}{apply local contour stylization}{boolean}{false}{Apply polynomial f0 contour stylization in local segments as e.g. AGs.}
\copar{navigate:do\_styl\_rhy\_en}{extract energy rhythm features}{boolean}{false}{apply DCT analyses on energy contour within user-defined segments and calculate the influence of events on the contour, in terms of the relative weight of DCT coefficients}
\copar{navigate:do\_styl\_rhy\_f0}{extract f0 rhythm features}{boolean}{false}{apply DCT analyses on f0 contour within user-defined segments and calculate the influence of events on the contour, in terms of the relative weight of DCT coefficients}
\copar{navigate:do\_styl\_voice}{extract voice quality features}{boolean}{false}{extract jitter and shimmer}
\copar{navigate:from\_scratch}{start from scratch}{boolean}{false}{If true, all configurations and analyses results in the pickle file are overwritten.}
\copar{navigate:overwrite\_config}{overwrite stored configurations}{boolean}{false}{If true, the configuration stored in the pickle file is overwritten by the current user-defined setting. Useful, if e.g. selected analysis steps should be repeated by different preprocessing settings.}

There are the following dependencies among the processing steps:

\begin{itemize}
\item all \myRef{do\_styl*} steps require preceding \myRef{do\_preproc}
\item \myRef{do\_styl\_loc} requires preceding \myRef{do\_styl\_glob}
\item \myRef{do\_styl\_bnd\*} requires preceding
  \myRef{do\_styl\_glob} if the boundary features are to be extracted
  from the f0 residuals.
\item all \myRef{do\_clst*} steps require a preceding \myRef{do\_styl*}
  step of the same type (\myRef{loc} or \myRef{glob})
\end{itemize}


If the preprocessing step \myRef{navigate:do\_preproc} is repeated, all
already extracted features are deleted since the updated preprocessing
configuration might lead to different stylization results. Thus by
repeating this step the user needs to redo all subsequent
stylizations.

\subsection{Directories, tiers, grouping}
\label{sec:dtg}
\copar{fsys:annot:dir}{annotation file directory}{string}{}{Can be nested. Depending on the task, audio, f0, and annotation files are obligatory or not. All obligatory directories must contain the same number of files in the same order. Optimally, same order is guaranteed using the same file name stem for corresponding audio, f0, and annotation files. However, this is not required.}
\copar{fsys:annot:ext}{annotation file extension}{string}{TextGrid, xml}{no default}
\copar{fsys:annot:typ}{annotation file type}{string}{TextGrid, xml}{Currently, only TextGrid and xml (see section \ref{sec:in_annot}) are supported. No default.}
\copar{fsys:aud:dir}{audio file directory}{string}{}{Can be nested. Depending on the task, audio, f0, and annotation files are obligatory or not. All obligatory directories must contain the same number of files in the same order. Optimally, same order is guaranteed using the same file name stem for corresponding audio, f0, and annotation files. However, this is not required.}
\copar{fsys:aud:ext}{audio file extension}{string}{}{Only files with this extension are collected from the directory.}
\copar{fsys:aud:typ}{audio file mimetype}{string}{wav}{currently only {\em wav} supported}
\copar{fsys:augment:chunk:tier\_out\_stm}{tier name stem of chunking output}{string}{chunk}{To the name stem the channel index will be added ({\bf also for mono files!}). E.g. given a stereo file and \myRef{fsys:augment:chunk:tier\_out\_stm=CHUNK}, the two segment tiers {\em CHUNK\_1} and {\em CHUNK\_2} will be generated for channel 1 and 2, respectively.}
\copar{fsys:augment:glob:tier\_out\_stm}{phrasing output tier}{string}{glob}{tier name stem of phrasing output. To the name stem the channel index will be added (also for mono files!). E.g. given a stereo file and \myRef{fsys:augment:glob:tier\_out\_stm=''IP''}, the two segment tiers {\em IP\_1} and {\em IP\_2} will be generated for channel 1 and 2, respectively.}
\copar{fsys:augment:glob:tier\_parent}{parent tier for prosodic phrase extraction}{string or list of strings}{{\bf fsys:augment:chunk:tier\_out\_stm}}{Segment tiers defining the superordinate domain for overall trend measurement from which the pre- and post- candidate-boundary segment deviate. This field can contain a single string (a single tier for mono files or any \myRef{fsys:augment:*:tier\_out\_stm} value which will be expanded by the channel index). The user can also explicitly specify multiple tier names in a list, if several channels are to be processed and the tier names cannot be derived from any \myRef{fsys:augment:*:tier\_out\_stm}. For segment tiers only.}
\copar{fsys:augment:glob:tier}{The tier in which to look for the prosodic boundary candidates.}{{\bf fsys:augment:syl:tier\_out\_stm + '\_bnd'}}{string or list of strings}{This field can contain a single string (a single tier for mono files or any \myRef{fsys:augment:*:tier\_out\_stm} value which will be expanded by the channel index and the syllable boundary infix). The user can also explicitly specify multiple tier names in a list, if several channels are to be processed and the tier names cannot be derived from any \myRef{fsys:augment:*:tier\_out\_stm}. Tiers can be of segment or event type. Default is the the {\em \_bnd}-output of \myRef{fsys:augment:syl:tier\_out\_stm}. Note that treating all syllable boundaries as phrase boundary candidates may result in prosodic boundaries within words. Thus a word segmentation tier is strongly recommended.}
\copar{fsys:augment:loc:tier\_acc}{Pitch accent extraction event tier}{string or list of strings}{$[$ $]$}{Pitch accent candidate time stamps, e.g. syllable nucleus midpoints. This field can contain a single string (a single tier for mono files or \myRef{fsys:augment:syl:tier\_out\_stm} which will be expanded by the channel index). The user can also explicitly specify multiple tier names in a list, if several channels are to be processed and the tier names cannot be derived from \myRef{fsys:augment:syl:tier\_out\_stm}. For event tiers only. Field can be empty, but at least one of \myRef{fsys:augment:loc:tier\_ag} and \myRef{fsys:augment:loc:tier\_acc} needs to be specified. If only \myRef{fsys:augment:loc:tier\_ag}: analysis within segment; if only \myRef{fsys:augment:loc:tier\_acc}: analysis within symmetric window  of length \myRef{preproc:point\_win} centered on the time stamp; if both: analysis within {\em ag} segment, time normalization so that 0 position is at {\em acc} time stamp within {\em ag}.}
\copar{fsys:augment:loc:tier\_ag}{pitch accent extraction segment tier}{string or list of strings}{$[$ $]$}{Tier with segments that are potential accent groups segment domain. This field can contain a single string for mono files or a list of strings for more channels. Tiers can be of segment type only. Field can be empty, but at least one of \myRef{fsys:augment:loc:tier\_ag} and \myRef{fsys:augment:loc:tier\_acc} needs to be specified. If only \myRef{fsys:augment:loc:tier\_ag}: analysis within segment; if only \myRef{fsys:augment:loc:tier\_acc}: analysis within symmetric window of length \myRef{preproc:point\_win} centered on the time stamp; if both: analysis within {\em ag} segment, time normalization so that 0 position is at {\em acc} time stamp within {\em ag}.}
\copar{fsys:augment:loc:tier\_out\_stm}{accent output tier name stem}{string}{acc}{To the name stem the channel index will be added (also for mono files!). E.g. given a stereo file and \\\myRef{fsys:augment:loc:tier\_out\_stm=''ACC''}, the two event tiers {\em ACC\_1} and {\em ACC\_2} will be generated for channel 1 and 2, respectively.}
\copar{fsys:augment:loc:tier\_parent}{name of parent tier for pitch accent candidates}{string or list of strings}{$[$ $]$}{This parent tier contains segments of a superordinate domain with respect to which the deviation of the accent candidate segments or time stamps is calculated. This might be global segments or chunks. Fallback is file-level. Must be segment tiers. This field can contain a single string (a single tier for mono files or any \myRef{fsys:augment:*:tier\_out\_stm} value which will be expanded by the channel index). The user can also explicitly specify multiple tier names in a list, if several channels are to be processed and the tier names cannot be derived from any \myRef{fsys:augment:*:tier\_out\_stm}. Tiers can be segment tiers only.}
\copar{fsys:augment:syl:tier\_out\_stm}{tier name stem of syllable nucleus and boundary output}{string}{syl}{To the name stem the channel index will be added (also for mono files!). Syllable boundary tiers are further marked by the infix {\em bnd}. E.g. given a stereo file and \myRef{fsys:augment:syl:tier\_out\_stm=''SYL''}, the four event tiers {\em SYL\_1, SYL\_bnd\_1} and {\em SYL\_2, SYL\_bnd\_2} will be generated for syllable nuclei and boundaries and for channel 1 and 2, respectively.}
\copar{fsys:augment:syl:tier\_parent}{parent tier for syllable nucleus extraction}{string or list of strings}{{\bf chunk}}{The parent tier defines the boundaries over which the reference window for relative energy calculation must not cross. Fallback is file level. This field can contain a single string (a single tier for mono files or any \myRef{fsys:augment:*:tier\_out\_stm} value which will be expanded by the channel index). The user can also explicitly specify multiple tier names in a list, if several channels are to be processed and the tier names cannot be derived from any \myRef{fsys:augment:*:tier\_out\_stm}.}
\copar{fsys:bnd:tier}{boundary tier names}{string or list of strings}{$[$ $]$}{each channel can contain several tiers to be analyzed. Segment or event tiers. For segment tiers the boundary between adjacent segments is parameterized, and for point tiers, the boundary at time stamps.}
\copar{fsys:channel:myTiername+}{channel index for each relevant tier name in the annotation file}{int}{myChannelIdx}{For augmentation output tiers this configuration branch is generated automatically.}
\copar{fsys:chunk:tier}{chunk tier names}{string or list of strings}{$[$ $]$}{one item for each channel. In case of multiple channels and single string, this string (e.g. ``chunk'') is expanded to ``chunk\_1'', ``chunk\_2'' \ldots for each available channel index. If chunk tiers specified, their segments' boundaries are not crossed by analysis and normalization windows for most feature sets. For the {\em bnd\_trend} feature set pre- and post-boundary segments are limited by the start and endpoint of the superordinate chunk if \myRef{styl:bnd:cross\_chunk} set to true.}
\copar{fsys:export:csv}{output csv tables}{boolean}{true}{If true, for each extracted feature set a csv file is outputted together with a code template file to read the table in R. The file names are concatenated by \myRef{fsys:export:stm} and the name of the feature set.}
\copar{fsys:export:dir}{output directory}{string}{}{Directory in which all csv tables, the log file, and the pickle file are stored.}
\copar{fsys:export:f0\_preproc}{output preprocessed f0 contours}{boolean}{false}{If true, preprocessed f0 values are outputted for each input f0 file. The output format is as specified in section \ref{sec:in_f0}. The output is stored in the subdirectory {\em f0\_preproc\/} below the directory \myRef{fsys:export:dir}.}
\copar{fsys:export:f0\_residual}{output residual f0 contours}{boolean}{false}{If true, residual f0 contours after register removal are outputted for each input f0 file. The output format is as specified in section \ref{sec:in_f0}. The output is stored in the subdirectory {\em f0\_residual\/} below the directory \myRef{fsys:export:dir}.}
\copar{fsys:export:f0\_resyn}{output resynthesized f0 contours}{boolean}{false}{If true, the resynthesized f0 contours as a superposition of global and local contour shapes are outputted for each input f0 file. The output format is as specified in section \ref{sec:in_f0}. The output is stored in the subdirectory {\em f0\_resyn\/} below the directory \myRef{fsys:export:dir}.}
\copar{fsys:export:fullpath}{whether or not to write the full path to the csv tables into the R code template files}{boolean}{false}{If true, the full path to the csv tables is written into the R code. 0 is recommended in case the data is shared and further processed at different locations.}
\copar{fsys:export:sep}{table column separator}{string}{{\bf ,}}{column separator for csv output tables.}
\copar{fsys:export:stm}{output file name stem}{string}{{\bf copasul}}{Same file name stem for all csv files, the log file, and the pickle file.}
\copar{fsys:export:summary}{output file/channel summary statistics}{boolean}{false}{If true, mean and variance values are calculated for all continuous-valued features outputted in the feature-set related csv files per file and analysis tier. For categorical features unigram entropies are calculated. A \myRef{fsys:export:stm}.summary.csv file is outputted together with an R code template file to read the table in R.}
\copar{fsys:f0:dir}{f0 file directory}{string}{}{Can be nested. Depending on the task, audio, f0, and annotation files are obligatory or not. All obligatory directories must contain the same number of files in the same order. Optimally, same order is guaranteed using the same file name stem for corresponding audio, f0, and annotation files. However, this is not required.}
\copar{fsys:f0:ext}{F0 file extension}{string}{}{only files with this extension are collected from the directory}
\copar{fsys:f0:typ}{}{string}{tab}{Currently only {\em tab} supported.}
\copar{fsys:glob:tier}{global segment tier names}{string or list of strings}{$[$ $]$}{analysis tiers for global segment, only one per each channel supported, so that global and local segments can be assigned to each other. If taken over from \myRef{fsys:augment:*:tier\_out\_stm}, the names must be extended by the corresponding channel index, e.g. IP\_1 etc, see \myRef{fsys:augment:*:tier\_out\_stm}. Segment or event tier. Events are considered to be right boundaries of segments and are expanded accordingly to segments.}
\copar{fsys:gnl\_en:tier}{Tiers for standard energy variable extraction.}{string or list of strings}{$[$ $]$}{More than one tier per channel supported. Segment or event tiers. Events are expanded to segments by \myRef{preproc:point\_win}.}
\copar{fsys:gnl\_f0:tier}{Tiers for standard f0 variable extraction}{string or list of strings}{$[$ $]$}{More than one tier per channel supported. Segment or event tiers. Events are expanded to segments by \myRef{preproc:point\_win}.}
\copar{fsys:grp:lab}{grouping labels with values derived from file names}{list of strings}{$[$ $]$}{Labels of file-name derived grouping. Non-relevant file parts are indicated by empty strings ''. E.g. given the f0 filename stem a\_b\_2. Let's say, ``a'' represents the speaker ID, ``b'' is not relevant for the current analysis, and ``2'' represents the stimulus ID. Then set \myRef{fsys:grp:src=f0}, \myRef{fsys:grp:sep=\_}, and \myRef{fsys:grp:lab=$[$'spk','','stim'$]$}. The output csv tables then contain two additional grouping columns {\em grp\_spk} and {\em grp\_stim} with values derived from the file names (in this case ``a'' and ``2''). Note that all grouping values are treated as strings.}
\copar{fsys:grp:sep}{file name split pattern}{string}{}{How to split the file name to access the grouping values. The string is interpreted as a regular expression. Thus predefined characters as the dot need to be protected! Thus if file parts are separated by a dot set this option to ``\textbackslash\textbackslash.''. If fileparts are separated by more than one symbol, e.g. dot and underscore, use ``(\_$|$\textbackslash.)''.}
\copar{fsys:grp:src}{grouping source}{string}{{\bf f0}, annot, aud}{from which file type to derive the file name based grouping}
\copar{fsys:label:chunk}{chunk label}{string}{x}{will be used by automatic chunking}
\copar{fsys:label:pau}{pause label}{string}{$<$P$>$}{in annotation files, segments labeled by this symbol are treated as pauses and are not analyzed. For boundary feature extraction these segments define the pause length feature between the preceding and following segment. Note, that this symbol as a pause identifier must be uniform over all analyzed tiers. In Praat TextGrids also not labeled segments are considered as pauses.}
\copar{fsys:label:syl}{syllable label}{string}{x}{will be used by automatic syllable extraction}
\copar{fsys:loc:tier\_acc}{local event tier names}{string or list of strings}{$[$ $]$}{tier (one for each channel) defining pitch accent time stamps. Event tiers only. Field can be empty, but at least one of \myRef{fsys:loc:tier\_ag} and \myRef{fsys:loc:tier\_acc} needs to be specified. If only \myRef{fsys:loc:tier\_ag}: analysis within segment; if only \myRef{fsys:loc:tier\_acc}: analysis within symmetric window of length \myRef{preproc:point\_win} centered on the time stamp; if both: analysis within {\em ag} segment, time normalization so that 0 position is at {\em acc} time stamp within {\em ag}.}
\copar{fsys:loc:tier\_ag}{local segment tier names}{string or list of strings}{$[$ $]$}{tier (one for each channel) defining accent group-like units. Segment tiers only. Field can be empty, but at least one of \myRef{fsys:loc:tier\_ag} and \myRef{fsys:loc:tier\_acc} needs to be specified. If only \myRef{fsys:loc:tier\_ag}: analysis within segment; if only \myRef{fsys:loc:tier\_acc}: analysis within symmetric window of length \myRef{preproc:point\_win} centered on the time stamp; if both: analysis within {\em ag} segment, time normalization so that 0 position is at {\em acc} time stamp within {\em ag}.}
\copar{fsys:pho:tier}{name of tier with phonetic segments}{string or list of strings}{$[]$}{one tier per channel. Used for feature extraction in prosodic boundary and accent localization.}
\copar{fsys:pho:vow}{vowel pattern}{string}{$[$AEIOUYaeiouy29\{$]$}{to identify vowel segments in \myRef{fsys:pho:tier}. Is interpreted as a regular expression.}
\copar{fsys:pic:dir}{directory for plotting output}{string}{}{directory for the png files generated by plotting.}
\copar{fsys:pic:stm}{file name stem of the plot files}{string}{{\bf copasul}}{}
\copar{fsys:pulse:dir}{Pulse file directory}{string}{}{Can be nested. Only for extracting voice quality features pulse files are obligatory. All obligatory directories must contain the same number of files in the same order. Optimally, same order is guaranteed using the same file name stem for corresponding audio, f0, pulse, and annotation files. However, this is not required.}
\copar{fsys:pulse:ext}{Pulse file extension}{string}{}{only files with this extension are collected from the directory}
\copar{fsys:pulse:typ}{}{string}{tab}{Currently only {\em tab} supported.}
\copar{fsys:rhy\_en:tier}{Tiers for energy rhythm extraction}{string or list of strings}{$[$ $]$}{More than one tier per channel supported. Segment or event tiers. Events are expanded to segments by \myRef{preproc:point\_win}.}
\copar{fsys:rhy\_en:tier\_rate}{Tiers containing units whose rate is to be calculated within each segment of the \myRef{fsys:rhy\_f0:tier} tiers}{string or list of strings}{$[$ $]$}{More than one tier per channel supported. Segment or event tiers.}
\copar{fsys:rhy\_f0:tier}{Tiers for f0 rhythm extraction}{string or list of strings}{$[$ $]$}{More than one tier per channel supported. Segment or event tiers. Events are expanded to segments by \myRef{preproc:point\_win}.}
\copar{fsys:rhy\_f0:tier\_rate}{Tiers containing units whose rate is to be calculated within each segment of the \myRef{fsys:rhy\_f0:tier} tiers}{string or list of strings}{$[$ $]$}{More than one tier per channel supported. Segment or event tiers.}
\copar{fsys:voice:tier}{Tiers for voice quality extraction}{string or list of strings}{$[$ $]$}{More than one tier per channel supported. Segment or event tiers. Events are expanded to segments by \myRef{preproc:point\_win}.}

\subsection{F0 preprocessing, windowing}
\copar{preproc:base\_prct}{Percentile below which base value for semitone transform is calculated}{\mybndfloat}{5}{Base value for semitone transform is defined as median of the values below the specified percentile. If set to 0, the base value will be set to 1, i.e. the semitone transform is carried out without normalization.}
\copar{preproc:base\_prct\_grp:myChannelIndex}{Grouping variable for which for each of its levels a base value for f0 semitone transform is calculated}{string}{' '}{Indicates for each channel index, which grouping variable is relevant. The grouping variable must be extractable from the file name as specified in \myRef{fsys:grp}. E.g. \myRef{preproc:base\_prct\_grp:1}{\em =spkId} requires a {\em spkId} element in the list of \myRef{fsys:grp:lab}. Channel indices must be written in quotation marks as strings.}
\copar{preproc:loc\_align}{Robust treatment of local segments to which more than one center is assigned in the annotation.}{string}{{\bf skip}, left, right}{{\em skip} -- such local segments are skipped; {\em left} -- the first center is kept; {\em right} -- the last center is kept.}
\copar{preproc:loc\_sync}{Extract {\em gnl\_*} and {\em rhy\_*} features only at locations where {\em loc} features can be obtained.}{boolean}{false}{Due to the strict hierarchy principle and to window length constraints it is not always possible to extract loc features at any location where gnl and rhy features can be obtained. If the user is interested only in locations where all these feature sets are available, so that the corresponding feature matrices can be concatenated, this option should be set to true.}
\copar{preproc:nrm\_win}{normalization window length (in sec)}{float}{0.6}{length of the normalization window. For feature sets {\em gnl\_*} all mean, max, std values derived in the analysis window are normalized within longer time window which length is defined by this parameter. If segments to be analyzed are longer than the normalization window, this window is set equal to the analyzed segment. \myRef{nrm\_win} can also be individually set for each of the feature sets {\em loc, gnl\_f0, gnl\_en, rhy\_f0, rhy\_en} (see section \ref{sec:feat}) by specifying \myRef{preproc:myFeatureSet:nrm\_win}.}
\copar{preproc:out:f}{outlier definition factor}{float}{2}{identifies non-zero f0 values as outliers, that deviate more than this factor times dispersion from the mean value. If \myRef{preproc:out:m=mean}, the mean value is given by the arithmetic mean and the dispersion by the standard deviation. If \myRef{preproc:out:m=median}, the mean value is given by the median and the dispersion by the inter quartile range. If \myRef{preproc:out:m=fence}, instead of the mean value the first and third quartiles are used as references and dispersion is given by the interquartile range ({\em Tukey's fences}).}
\copar{preproc:out:m}{reference value definition for outlier identification}{string}{{\bf mean}, median, fence}{Specifies definition of mean/fence and dispersion, see \myRef{preproc:out:f} for details.}
\copar{preproc:point\_win}{window length to transform events to segments (in sec)}{float}{0.3}{The extraction of the feature sets {\em glb\_*, rhy\_*, glob, loc} is based on segments. For event tier input, segments are obtained by centering a window of this length on the time stamps. \myRef{point\_win} can also be individually set for each of the feature sets {\em loc, gnl\_f0, gnl\_en, rhy\_f0, rhy\_en} (see section \ref{sec:feat}) by specifying \myRef{preproc:myFeatureSet:point\_win}.}
\copar{preproc:smooth:mtd}{F0 smoothing method}{string}{{\bf sgolay}, med}{Savitzky-Golay or median filtering of f0 contour. Median yields stronger smoothing, Savitzky-Golay performs better in keeping local minima and maxima at their place.}
\copar{preproc:smooth:ord}{polynomial order of smoothing method}{integer}{3}{relevant for \myRef{preproc:smooth:mtd=sgolay} only.}
\copar{preproc:smooth:win}{smoothing window length (in f0 sample indices)}{int}{7}{The longer the smoothing window, the more smooth the f0 contours.}
\copar{preproc:st}{Hertz to semitone conversion}{boolean}{true}{If true, transformed to semitones.}

\subsection{Augmentation: Chunking}
\copar{augment:chunk:e\_rel}{proportion of reference energy below which a pause is assumed}{float}{0.0767}{a pause is indicated, if the energy in the analysis window is below this factor times the energy in the longer reference window.}
\copar{augment:chunk:fbnd}{assume pause at beginning and end of file}{boolean}{true}{If set to true, forced pause detection at file start and end. These pauses are subtracted from \myRef{augment:chunk:n} if set.}
\copar{augment:chunk:flt:btype}{filter type}{string}{{\bf low}, high, band}{Butterworth filter type to filter the signal for pause detection. Recommended: {\em low}.}
\copar{augment:chunk:flt:f}{filter cutoff frequencies (in Hz)}{float or list of floats}{{\bf 8000}}{For \myRef{augment:chunk:flt:btype=low, high} a single cut-off frequency is expected; for {\em band} a 2-element list of lower and upper cutoff frequency.}
\copar{augment:chunk:flt:ord}{filter order}{int}{5}{Butterworth filter order.}
\copar{augment:chunk:l\_ref}{reference window length for pause detection (in sec)}{int}{5}{Energy in analysis window of length \myRef{augment:chunk:l} is compared against the energy within the reference window. Same midpoint as analysis window.}
\copar{augment:chunk:l}{length of the analysis window (in sec)}{float}{0.1524}{analysis window for which is to be decided, whether or not it is (part of) a pause.}
\copar{augment:chunk:margin}{silence margin at chunk start and end (in sec)}{float}{0}{chunks are extended by this amount on both sides.}
\copar{augment:chunk:min\_chunk\_l}{minimum chunk length (in sec)}{float}{0.3}{shorter chunks are merged}
\copar{augment:chunk:min\_pau\_l}{minimum pause length (in sec)}{float}{0.3}{shorter pauses are ignored.}
\copar{augment:chunk:n}{pre-specified number of pauses $[$sic!$]$}{integer}{-1}{In this implementation chunks are defined as interpausal units and thus depend on pause detection. If set to -1, no pre-specified pause number.}

\subsection{Augmentation: Syllable nucleus detection}
\copar{augment:syl:d\_min}{minimum distance between subsequent syllable nuclei (in sec)}{float}{0.05}{If 2 detected nuclei are closer than this distance the weaker candidate is discarded.}
\copar{augment:syl:e\_min}{minimum energy factor relative to entire file}{float}{0.16}{For a syllable nucleus the RMS energy in the analysis window must be above this factor times the energy in the entire file.}
\copar{augment:syl:e\_rel}{minimum energy factor relative to reference window}{float}{1.07}{For a syllable nucleus the RMS energy in the analysis window must be above this factor times the energy in the reference window.}
\copar{augment:syl:flt:btype}{filter type}{string}{low, high, {\bf band}}{Butterworth filter type to filter the signal for syllable nucleus detection. Recommended: {\em band}.}
\copar{augment:syl:flt:f}{filter cutoff frequencies (in Hz)}{float or list of floats}{[ 200 4000]}{For \myRef{augment:syl:flt:btype=low, high} a single cut-off frequency is expected; for {\em band} a 2-element list of lower and upper cutoff frequency.}
\copar{augment:syl:flt:ord}{filter order}{int}{5}{Butterworth filter order.}
\copar{augment:syl:l\_ref}{reference window length for syllable detection (in sec)}{float}{0.15}{Energy in analysis window with same midpoint is compared against the energy within the reference window.}
\copar{augment:syl:l}{analysis window length (in sec)}{float}{0.08}{length of window within energy is calculated. Same midpoint as reference window.}

\subsection{Augmentation: Prosodic boundary detection}
\copar{augment:glob:cntr\_mtd}{how to define cluster centroids}{string}{{\bf seed\_prct}, seed\_kmeans, split}{{\em seed\_*}: initialize clustering by bootstrapped seed centroids. {\em seed\_prct:} single-pass clustering of the boundary candidates by their distance to these centroids. Distance values to the no-boundary seed above a specified percentile \myRef{augment:glob:prct} indicate boundaries. {\em seed\_kmeans:} kmeans clustering initialized by the seed centroids (gives a more balanced amount of boundary/no boundary cases than {\em seed\_prct}). {\em split}: centroids are derived by splitting each column in the feature matrix at the percentile \myRef{augment:glob:prct}; the boundary centroid is defined by the median of the values above the splitpoint, the no-boundary centroid by the median of the values below; items are then assigned to the nearest centroid in a single pass. Depending on \myRef{augment:glob:unit} clustering is either carried out separately within each file and each channel, or over the entire dataset. Fallback: if cluster centroids cannot be bootstrapped, this parameter's value is changed to {\em split}.}
\copar{augment:glob:heuristics}{heuristic macro settings}{string}{ORT}{Only {\em ORT} supported. {\em ORT} assumes a word segmentation tier for prosodic boundary prediction and rejects boundaries after too short and thus probably function words ($<0.1s$). Not necessarily meaningful for any language.}
\copar{augment:glob:measure}{feature values, or deltas}{string}{{\bf abs}, delta, abs+delta}{Which values $v$ to put in the feature matrix ($i$=time index): {\em abs}: feature values $v[i]$; {\em delta}: feature deltas $v[i]-v[i-1]$; {\em abs+delta}: both}
\copar{augment:glob:min\_l}{minimum inter-boundary distance (in sec)}{float}{0.5}{If 2 detected boundaries are closer than this value, only the stronger one will be kept. This distance is also used in bootstrapping boundary and no-boundary centroids as described in section \ref{sec:augbnd}.}
\copar{augment:glob:prct}{percentile of cluster splitpoint}{\mybndfloat}{95}{Splitpoint definition for clustering in terms of a percentile value. The higher the fewer boundaries will be detected. For \myRef{augment:glob:cntr\_mtd=split} the percentile refers to the feature values, for \myRef{augment:glob:cntr\_mtd=seed\_prct}, it refers to the distance to the no-boundary seed centroid.}
\copar{augment:glob:seed}{random seed}{int}{42}{random state for KMeans for clustering results reproducibility.}
\copar{augment:glob:unit}{derive centroids separately for each file or over entire data set}{string}{{\bf batch}, file}{batch mode recommended for corpora containing lots of short recordings, within which centroids cannot reliably be extracted.}
\copar{augment:glob:wgt:myBndFeatset+:myRegister+:myFeat+}{user defined feature weights}{float}{1}{create one config branch for each selected boundary feature and assign a weight. Only boundary features supported. The weight becomes a dummy in case of \myRef{augment:glob:wgt\_mtd} is not {\em user}. However, the branches must be specified in order to mark which features to be used for boundary prediction. {\em myBndFeatset $\in$ \{ std, win, trend\}, myRegister $\in$ \{bl, ml, tl, rng\}, myFeat $\in$ \{r, rms, rms\_pre, \ldots\}}. The branches must correspond to branches in the sub-dictionary \myRef{copa:data:myFileIdx:myChannelIdx:bnd:myTierNameIndex:myBoundaryIndex} (see section \ref{sec:dict}).\\E.g. \myRef{copa:data:myFileIdx:myChannelIdx:bnd:myTierNameIndex:myBoundaryIndex:win:bl:r} is addressed by\\ \myRef{augment:glob:wgt:win:bl:r}.}
\copar{augment:glob:wgt:pho}{use/weight normalized vowel length as feature}{float}{1}{only compliant with \myRef{augment:glob:unit=batch}.}
\copar{augment:glob:wgt\_mtd}{feature weighting method}{string}{{\bf silhouette}, correlation, user}{For {\em silhouette} an initial clustering is carried out, and for each feature its weight is then defined by its cluster-separating power. For {\em correlation} weights are defined for each feature by its correlation to the feature vector medians. For {\em user}, the weights specified in the \myRef{augment:glob:wgt:myBndFeatset+:myRegister+:myFeat+} branches are taken.}

\subsection{Augmentation: Pitch accent detection}
\copar{augment:loc:acc\_select}{which syllable within a segment to select}{string}{{\bf max}, left, right}{Choose the accent position among all time stamps in \myRef{augment:loc:tier\_acc} that are in the same segment of \myRef{fsys:augment:loc:tier\_ag}. {\em max:} the most prominent one; {\em left, right}: accent first/last syllable, which might be useful if \myRef{fsys:augment:loc:tier\_ag} contains word segments, and word stress is fixed.}
\copar{augment:loc:ag\_select}{which segments to select for accentuation}{string}{{\bf max}, all}{{\em all:} assign an accent to each segment in \myRef{fsys:augment:loc:tier\_ag}; {\em max:} assign accents to the most prominent segments only.}
\copar{augment:loc:cntr\_mtd}{how to define cluster centroids}{string}{{\bf seed\_prct}, seed\_kmeans, split}{{\em seed\_*}: initialize clustering by bootstrapped seed centroids. {\em seed\_prct:} single-pass clustering of the accent candidates by their distance to these centroids. Distance values to the no-accent seed above a specified percentile \myRef{augment:loc:prct} indicate accents. {\em seed\_kmeans:} kmeans clustering initialized by the seed centroids (gives a more balanced amount of accent/no-accent cases than {\em seed\_prct}). {\em split}: centroids are derived by splitting each column in the feature matrix at the percentile \myRef{augment:glob:prct}; the accent centroid is defined by the median of the values above the splitpoint, the no-accent centroid by the median of the values below; items are then assigned to the nearest centroid in a single pass. Depending on \myRef{augment:loc:unit} clustering is carried out either separately within each file and each channel, or over the entire dataset. Fallback: if cluster centroids cannot be bootstrapped, this parameter's value is changed to {\em split}.}
\copar{augment:loc:heuristics}{heuristic macro settings}{string}{ORT}{only {\em ORT} supported. {\em ORT} assumes a word segmentation tier for accent extraction. Short words (see \myRef{augment:loc:max\_l\_na}) will be treated as non-accent seeds, long words (see \myRef{augment:loc:min\_l\_a}) as accent seeds.}
\copar{augment:loc:max\_l\_na}{maximum length of definitely non-accented words (in sec)}{float}{0.1}{from words below that length the {\em non-accented} seed centroid is derived}
\copar{augment:loc:measure}{feature values, or deltas}{string}{{\bf abs}, delta, abs+delta}{Which values $v$ to put in the feature matrix ($i$=time index): {\em abs}: feature values $v[i]$; {\em delta}: feature deltas $v[i]-v[i-1]$; {\em abs+delta}: both}
\copar{augment:loc:min\_l\_a}{minimum length of definitely accented words (in sec)}{float}{0.6}{from words above that length the {\em accented} seed centroid is derived}
\copar{augment:loc:min\_l}{minimum inter-accent distance (in sec)}{float}{0.2}{If 2 detected accents are closer than this value, only the more prominent one will be kept.}
\copar{augment:loc:prct}{percentile of cluster splitpoint}{\mybndfloat}{90}{Splitpoint definition for clustering in terms of a percentile value. The higher the fewer accents will be detected. For \myRef{augment:loc:cntr\_mtd=split} the percentile refers to the feature values, for \myRef{augment:loc:cntr\_mtd=seed\_prct}, it refers to the distance to the no-accent seed centroid.}
\copar{augment:loc:seed}{random seed}{int}{42}{random state for KMeans for clustering results reproducibility.}
\copar{augment:loc:unit}{derive centroids separately for each file or over entire data set}{string}{{\bf batch}, file}{batch mode recommended for corpora containing lots of short recordings, within which centroids cannot reliably be extracted.}
\copar{augment:loc:wgt:myFeatset+:\ldots}{user defined feature weights}{float}{1}{create one config branch for each selected prominence feature and assign a weight. {\em myFeatset $\in$ \{acc, gst, gnl\_f0, gnl\_en\}}. The weight becomes a dummy in case of \myRef{augment:loc:wgt\_mtd} is not {\em user}. However, the branches must be specified in order to mark which features to be used for accent prediction. The branches must correspond to branches in the sub-dictionary \myRef{copa:data:myFileIdx:myChannelIdx:loc} (see section \ref{sec:dict}). E.g. \myRef{copa:data:myFileIdx:myChannelIdx:loc:gst:bl:rms} is addressed by \myRef{augment:loc:wgt:gst:bl:rms}. If the value at this branch is a list (e.g. the polynomial coefficients in \myRef{\ldots augment:loc:wgt:acc:c}) the weight can either be a scalar to weight all list elements equally or a list of same length as the value list, to individually weight each element. (Only) for polynomial coefficients absolute values are taken.}
\copar{augment:loc:wgt:pho}{use/weight normalized vowel length as feature}{float}{1}{only compliant with \myRef{augment:loc:unit=batch}.}
\copar{augment:loc:wgt\_mtd}{feature weighting method}{string}{{\bf silhouette}, correlation, user}{For {\em silhouette} an initial clustering is carried out, and for each feature its weight is then defined by its cluster-separating power. For {\em correlation} weights are defined for each feature by its correlation to the feature vector medians. For {\em user}, the weights specified in the \myRef{augment:loc:wgt:\ldots+} branches are taken.}

\subsection{Stylization: Global contours}
\copar{styl:glob:decl\_win}{window length for median calculation (in sec)}{float}{0.1}{Within each window a median each for the base-, mid-, and topline is derived.}
\copar{styl:glob:nrm:mtd}{time normalization method}{string}{{\bf minmax}}{for time normalization in global segment. Currently only {\em minmax} supported.}
\copar{styl:glob:nrm:rng}{normalized time range}{list of floats}{$[0,1]$}{normalized time of segment start and endpoint}
\copar{styl:glob:prct:bl}{percentile below which the baseline input medians are calculated}{\mybndfloat}{10}{A sequence of lower range medians is calculated along the f0 contour. The baseline is given by linear regression through this sequence.}
\copar{styl:glob:prct:tl}{percentile, above which the topline input medians are calculated}{\mybndfloat}{90}{A sequence of upper range medians is calculated along the f0 contour. The topline is given by linear regression through this sequence.}

\subsection{Stylization: Local contours}
\copar{styl:loc:nrm:mtd}{time normalization method}{string}{{\bf minmax}}{for time normalization in the local segment. Currently only {\em minmax} supported.}
\copar{styl:loc:nrm:rng}{normalized time range}{list of floats}{$[-1,1]$}{normalized time of segment start and endpoint. $[-1,1]$ is recommended to center the polynomial around 0.}
\copar{styl:loc:ord}{polynomial order}{int}{3}{Each coefficient will get its output column in the exported tables, thus the table size depends on this order.}

\subsection{Stylization: Register representation}
\copar{styl:register}{register definition for residual calculation}{string}{{\bf ml}, bl, tl, rng, none}{how to remove the global component from the f0 contour to get the residual the local contour is calculated on; {\em bl, ml, tl}: base-, mid- or topline subtraction; {\em rng} pointwise $[0$ $1]$ normalization of the f0 contour with respect to the base- and topline. Recommended: {\em ml, rng}. {\rm rng} normalizes for range declination (lower f0 amplitudes at the end of prosodic phrases).}

\subsection{Stylization: Boundaries}
\copar{styl:bnd:cross\_chunk}{stylization windows across chunks}{boolean}{true}{If set to true, the windows defined by \myRef{styl:bnd:win} can cross chunks, else they are limited by the current chunk's boundaries. If set to true for \myRef{do\_bnd\_trend}, lines are fitted from file start and till file end. Else, they are limited by the current chunk's boundaries.}
\copar{styl:bnd:decl\_win}{window length for median calculation (in sec)}{float}{0.1}{Within each window a median each for the base-, mid- and topline is derived.}
\copar{styl:bnd:nrm:mtd}{time normalization method}{string}{{\bf minmax}}{Only minmax supported.}
\copar{styl:bnd:nrm:rng}{normalized time range}{list of floats}{$[0,1]$}{to allow for comparisons independent of segment length, time is normalized to this range.}
\copar{styl:bnd:prct:bl}{percentile below which the baseline input medians are calculated}{\mybndfloat}{10}{A sequence of lower range medians is calculated along the f0 contour. The baseline is given by linear regression through this sequence.}
\copar{styl:bnd:prct:tl}{percentile, above which the topline input medians are calculated}{\mybndfloat}{90}{A sequence of upper range medians is calculated along the f0 contour. The topline is given by linear regression through this sequence.}
\copar{styl:bnd:residual}{use f0 residual}{boolean}{false}{measure discontinuity on (preprocessed) f0 contour or on its residual after register subtraction.}
\copar{styl:bnd:win}{window length (in sec)}{float}{1}{stylization window length for \myRef{navigate:do\_styl\_bnd\_win}}

\subsection{Stylization: General (energy) features}
\copar{styl:gnl:sb:alpha}{pre-emphasis factor or lower boundary frequency}{float}{0.95}{For pre-emphasis in the time domain for spectral balance calculation. $0\leq\alpha\leq1$: factor in $s'[i] = s[i]-\alpha \cdot s[i-1]$. $\alpha>1$: lower boundary frequency from which pre-emphasis should start. Will be internally converted to the factor in the above formula.}
\copar{styl:gnl:sb:btype}{filter type to restrict frequency window}{string}{{\bf none}, band, high, low}{Restrict frequency window for spectral balance calculation}
\copar{styl:gnl:sb:domain}{domain for spectral balance calculation}{string}{{\bf time}, freq}{Specifies whether spectral balance should be calculated in the time ('time') or frequency ('freq') domain.}
\copar{styl:gnl:sb:f}{filter cutoff frequencies (in Hz) for spectral balance calculation}{float or list of floats}{-1}{Specifies the upper cutoff frequency for a low-pass filter, the lower cutoff frequency for a high-pass filter, or both for a bandpass filter. See \myRef{styl:gnl:sb:btype}.}
\copar{styl:gnl:sb:win}{length (in sec) of central analysis window in analysed segment}{float}{-1}{To be set if coarticulatory influence on spectral balance calculation should be removed. If -1 the entire segment is used.}
\copar{styl:gnl:win}{window length to determine initial and final part of contour}{float}{0.3}{Length of window (in sec) for initial and final part of f0 or energy contour to calculate mean f0 pr energy quotients of these parts and the entire contour.}
\copar{styl:gnl\_en:alpha}{pre-emphasis factor}{float}{0.95}{Pre-emphasis is carried out in the time domain as follows: $s'[i] = s[i]-\alpha \cdot s[i-1]$.  {\bf DEPRECATED! NOW SPECIFIED BY styl:gnl\_en:sb:alpha}}
\copar{styl:gnl\_en:sts}{step size (in sec)}{float}{0.01}{Stepsize by which energy window is shifted.}
\copar{styl:gnl\_en:winparam}{window parameter}{string or int}{null}{Depends on \myRef{styl:gnl\_en:wintyp}; as required by {\em scipy.signal.get\_window()}.}
\copar{styl:gnl\_en:wintyp}{window type}{string}{{\bf hamming}, kaiser, \ldots}{All window types that are supported by {\em scipy.signal.get\_window()} can be used.}
\copar{styl:gnl\_en:win}{window length (in sec)}{float}{0.05}{Energy is calculated in terms of RMSD within windows of this length.}

\subsection{Stylization: F0 rhythm features}
\copar{styl:rhy\_f0:rhy:lb}{Lower frequency boundary of DCT coefficients (in Hz)}{float}{0.0}{Can be raised if low-frequency events should be ignored.}
\copar{styl:rhy\_f0:rhy:nsm}{number of spectral moments}{int}{3}{How many spectral moments to calculate from DCT analysis of f0 contour.}
\copar{styl:rhy\_f0:rhy:rmo}{remove DCT offset}{boolean}{false}{Remove first DCT coefficient.}
\copar{styl:rhy\_f0:rhy:ub}{upper frequency boundary of DCT coefficients (in Hz)}{float}{10}{Upper boundary of analyzed DCT spectrum (higher-frequency events assumed not to be influential for prosody).}
\copar{styl:rhy\_f0:rhy:wgt:rb}{rate band (in Hz)}{float}{1}{Frequency band around event frequency, within which the influence of the event in terms of absolute DCT coefficient values is integrated. E.g. for an event rate of 4 Hz and a rate band of 1 Hz the absolute values of the DCT coefficients between 3 and 5 Hz are summed up.}
\copar{styl:rhy\_f0:rhy:winparam}{window parameter}{string or int}{1}{depends on \myRef{styl:gnl\_en:wintyp}; as required by {\em scipy.signal.get\_window()}.}
\copar{styl:rhy\_f0:rhy:wintyp}{window type for DCT analysis}{string}{hamming, {\bf kaiser}, \ldots}{All window types that are supported by {\em scipy.signal.get\_window()} can be used.}

\subsection{Stylization: Energy rhythm features}
\copar{styl:rhy\_en:rhy:lb}{lower frequency boundary of DCT coefficients (in Hz)}{float}{0}{Can be raised if low-frequency events should be ignored.}
\copar{styl:rhy\_en:rhy:nsm}{number of spectral moments}{int}{3}{How many spectral moments to be calculated from DCT analysis of energy contour.}
\copar{styl:rhy\_en:rhy:rmo}{remove DCT offset}{boolean}{false}{Remove first DCT coefficient.}
\copar{styl:rhy\_en:rhy:ub}{upper frequency boundary of DCT coefficients (in Hz)}{float}{10}{Upper boundary of analyzed DCT spectrum (higher-frequency events assumed not to be influential for prosody).}
\copar{styl:rhy\_en:rhy:wgt:rb}{rate band (in Hz)}{float}{1}{Frequency band around event frequency, within which the influence of the event in terms of absolute DCT coefficient values is integrated. E.g. for an event rate of 4 Hz and a rate band of 1 Hz the absolute values of the DCT coefficients between 3 and 5 Hz are summed up.}
\copar{styl:rhy\_en:rhy:winparam}{DCT window parameter}{string or int}{1}{Depends on \myRef{styl:rhy\_en:wintyp}; as required by {\em scipy.signal.get\_window()}.}
\copar{styl:rhy\_en:rhy:wintyp}{window type for DCT}{string}{hamming, {\bf kaiser}, \ldots}{All window types that are supported by {\em scipy.signal.get\_window()}.}
\copar{styl:rhy\_en:sig:scale}{scale signal to maximum amplitude 1}{boolean}{true}{if set to true, the signal is scaled to its maximum amplitude. This is suggested especially if signals of different recording conditions are to be compared.}
\copar{styl:rhy\_en:sig:sts}{step size (in sec)}{float}{0.01}{Step size by which the energy window is shifted.}
\copar{styl:rhy\_en:sig:winparam}{window parameter}{string or int}{null}{Depends on \myRef{styl:rhy\_en:wintyp}; as required by {\em scipy.signal.get\_window()}.}
\copar{styl:rhy\_en:sig:wintyp}{window type of energy calculation}{string}{{\bf hamming}, kaiser, \ldots}{all window types that are supported by {\em scipy.signal.get\_window()}.}
\copar{styl:rhy\_en:sig:win}{window length (in sec)}{float}{0.05}{Energy is calculated in terms of RMSD within windows of this length.}

\subsection{Stylization: Voice quality features}
\copar{styl:voice:jit:fac\_max}{maximally allowed quotient of adjacent periods}{float}{1.3}{corresponds to Praat parameter Maximum period factor.}
\copar{styl:voice:jit:t\_max}{maximum period length in sec}{float}{0.02}{corresponds to Praat parameter Period ceiling.}
\copar{styl:voice:jit:t\_min}{minimum period length in sec}{float}{0.0001}{corresponds to Praat parameter Period floor.}

\subsection{Clustering: Global contours}
\copar{clst:glob:estimate\_bandwidth:n\_samples}{number of samples to estimate bandwidth}{integer}{1000}{Computationally expensive, high numbers will require long processing time.}
\copar{clst:glob:estimate\_bandwidth:quantile}{estimate\_bandwidth quantile parameter}{float}{0.3}{Lower values result in higher clusters numbers.}
\copar{clst:glob:kMeans:init}{initialization method of kmeans}{string}{meanShift}{All methods that are supported by {\em kMeans()} can be used. For {\em meanShift} the number of clusters does not need to be specified.}
\copar{clst:glob:kMeans:max\_iter}{kMeans: maximum number of iterations}{int}{300}{When to stop cluster re-adjustment, if not yet converged.}
\copar{clst:glob:kMeans:n\_cluster}{kMeans: predefined number of contour classes}{int}{3}{Irrelevant, if kmeans centroids are initialized by \myRef{clst:glob:kMeans:init=meanShift}.}
\copar{clst:glob:kMeans:n\_init}{number of initialization trials}{int}{10}{kMeans is repeated with different cluster initializations from which the best clustering result is kept.}
\copar{clst:glob:meanShift:bandwidth}{bandwidth parameter for meanShift cluster center initialization}{float}{0.0}{0.0 indicates, that the optimal bandwidth is internally calculated.}
\copar{clst:glob:meanShift:bin\_seeding}{bin seeding}{boolean}{false}{parameter for meanShift clustering.}
\copar{clst:glob:meanShift:min\_bin\_freq}{minimum number of items in each bin}{int}{1}{Parameter for meanShift clustering.}
\copar{clst:glob:mtd}{clustering method}{string}{{\bf meanShift}, kmeans}{No initial cluster number specification needed for {\em meanShift}.}
\copar{clst:glob:seed}{random seed}{int}{42}{random state for KMeans and bandwidth estimation for clustering results reproducibility.}

\subsection{Clustering: Local contours}
\copar{clst:loc:estimate\_bandwidth:n\_samples}{number of samples to estimate bandwidth}{int}{1000}{Computationally expensive, high numbers will require long processing time.}
\copar{clst:loc:estimate\_bandwidth:quantile}{estimate\_bandwidth quantile parameter}{float}{0.3}{Lower values result in higher clusters numbers.}
\copar{clst:loc:kMeans:init}{initialization method of kmeans}{string}{meanShift}{All methods that are supported by {\em kMeans()} can be used. For {\em meanShift} the number of clusters does not need to be specified.}
\copar{clst:loc:kMeans:max\_iter}{kMeans: maximum number of iterations}{int}{300}{When to stop cluster re-adjustment, if not yet converged.}
\copar{clst:loc:kMeans:n\_cluster}{kMeans: predefined number of contour classes}{int}{5}{Irrelevant, if kmeans centroids are initialized by \myRef{clst:glob:kMeans:init=meanShift}.}
\copar{clst:loc:kMeans:n\_init}{number of initialization trials}{int}{10}{kMeans is repeated with different cluster initializations from which the best clustering result is kept.}
\copar{clst:loc:meanShift:bandwidth}{bandwidth parameter for meanShift cluster center initialization}{float}{0.0}{0.0 indicates, that the optimal bandwidth is internally calculated.}
\copar{clst:loc:meanShift:bin\_seeding}{bin seeding}{boolean}{false}{parameter for meanShift clustering.}
\copar{clst:loc:meanShift:min\_bin\_freq}{minimum number of items in each bin}{int}{1}{Parameter for meanShift clustering.}
\copar{clst:loc:mtd}{clustering method}{string}{{\bf meanShift}, kmeans}{No initial cluster number specification needed for {\em meanShift}.}
\copar{clst:glob:seed}{random seed}{int}{42}{random state for KMeans and bandwidth estimation for clustering results reproducibility.}

\subsection{Plotting: Browsing}
\copar{plot:browse:grp}{plot for selected grouping values only}{dict}{empty}{This dict contains zero or more {\em myGroupingKey-myGroupingValue} pairs. Each {\em myGroupingKey} should match one of the strings in \myRef{fsys:grp:lab}. By this the user can select to plot images with (a combination of) certain grouping values only.}
\copar{plot:browse:save}{save plots according to \myRef{fsys:pic}}{boolean}{false}{Store png files in \myRef{fsys:pic:dir} with file name stem \myRef{fsys:pic:stm}.}
\copar{plot:browse:single\_plot:active}{switch on single plot mode}{boolean}{false}{switch on single plot mode if only one segment specified by file index, channel index, and segment index is to be plotted}
\copar{plot:browse:single\_plot:channel\_i}{channel index of selected segment}{integer}{0}{channel index of selected segment to be plotted}
\copar{plot:browse:single\_plot:file\_i}{file index of selected segment}{integer}{0}{file index of selected segment to be plotted}
\copar{plot:browse:single\_plot:segment\_i}{segment index of selected segment}{integer}{0}{segment index of selected segment to be plotted}
\copar{plot:browse:time}{when to do plotting}{string}{online, {\bf final}}{{\em online:} plot at stylization stage for immediate check of appropriateness of configurations. {\em final:} plot segment-wise from the finally stored results. Click on plot: next; press {\em return}: quit.}
\copar{plot:browse:type:clst:contours}{plot global and local intonation class centroids}{boolean}{false}{}
\copar{plot:browse:type:complex:gestalt}{plot local contour Gestalt stylization}{boolean}{false}{}
\copar{plot:browse:type:complex:superpos}{plot global and local contour superposition}{boolean}{false}{}
\copar{plot:browse:type:glob:decl}{plot global contour register stylization}{boolean}{false}{}
\copar{plot:browse:type:loc:acc}{plot local contour polynomial stylization}{boolean}{false}{}
\copar{plot:browse:type:loc:decl}{plot local contour register stylization}{boolean}{false}{}
\copar{plot:browse:type:complex:bnd}{plot boundary stylization}{boolean}{false}{}
\copar{plot:browse:type:complex:bnd\_win}{plot boundary stylization (fixed window)}{boolean}{false}{}
\copar{plot:browse:type:complex:bnd\_trend}{plot boundary stylization (trend)}{boolean}{false}{}
\copar{plot:browse:type:rhy\_en:rhy}{plot influence of rate tier events on DCT of energy contour in analysis tier}{boolean}{false}{}
\copar{plot:browse:type:rhy\_f0:rhy}{plot influence of rate tier events on DCT of f0 contour in analysis tier}{boolean}{false}{}
\copar{plot:browse:verbose}{display file, channel and segment index for each plot}{boolean}{false}{written to STDOUT}
\copar{plot:color}{plot in color (true) or black-white (false)}{boolean}{true}{}

\subsection{Plotting: Grouping}
\copar{plot:grp:grouping}{list of selected grouping variables from \myRef{fsys:grp:lab}}{list of strings}{$[$ $]$}{For each combination of grouping factor levels the stylization plot based on the respective parameter mean vector is stored as a png file in \myRef{fsys:pic:dir} with file name stem \myRef{fsys:pic:stm} and an infix expressing the respective factor level combination.}
\copar{plot:grp:save}{save plots according to \myRef{fsys:pic}}{boolean}{false}{Store png files in \myRef{fsys:pic:dir} with file name stem \myRef{fsys:pic:stm}. One file per group.}
\copar{plot:grp:type:glob:decl}{plot global contour declination centroid for each group}{boolean}{false}{Plots are not displayed but saved as png files to \myRef{fsys:pic}.}
\copar{plot:grp:type:loc:acc}{plot local contour polynomial shape centroid for each group}{boolean}{false}{Plots are not displayed but saved as png files to \myRef{fsys:pic}.}
\copar{plot:grp:type:loc:decl}{plot local contour declination centroid for each group}{boolean}{false}{Plots are not displayed but saved as png files to \myRef{fsys:pic}.}

\section{Output}
\label{sec:output}
\subsection{Table files}
\label{sec:otf}

If \myRef{fsys:export:csv} is set to true, for each feature set selected
by the \myRef{navigate:*} options a csv table file with
alphanumerically sorted columns is generated in
\myRef{config:fsys:export:dir}. The file name is the
underscore-concatenation of \myRef{config:fsys:export:stm} and the
feature set name. Extension is {\em csv}. Columns are separated by a
comma. The column titles correspond to the feature names given in the
tables in section \ref{sec:feat}, and each row corresponds to one
segment or event for which the features were extracted.  These feature
vectors are additionally linked to the data origin by the following
columns:

\begin{center}
  \begin{tabular}{l|l}
    {\bf name} & {\bf description} \\
    \hline
    ci & channel index (starting with 0) \\
    fi & file index (starting with 0) \\
    ii & item (segment or event) index (starting with 0) \\
    stm & annotation file name stem \\
    t\_on & time onset \\
    t\_off & time offset (same as t\_on for events) \\
    tier & tier name
  \end{tabular}
\end{center}

Inter-tier relations are provided by the following columns

\begin{center}
  \begin{tabular}{l|l}
    {\bf name} & {\bf description} \\
    \hline
    is\_init & initial position in a global segment \\
    is\_fin & final position in a global segment \\
    is\_init\_chunk & initial position in a chunk \\
    is\_fin\_chunk & final position in a chunk
  \end{tabular}
\end{center}

All columns contain the values {\em yes} and {\em no}. Medial position is
simply indicated by {\em is\_init=no} and {\em is\_fin=no}. These
columns can be used for data subsetting. As an example, let's assume
that boundary features were extracted between accent groups, and the
global segments correspond to intonation phrases. Then {\em is\_fin}
serves to hold apart IP-final and non-final boundaries. Equivalently,
phrase-final and non-final accents can be held apart. {\em
  is\_init\_chunk} and {\em is\_fin\_chunk} work the same on the chunk
level. If no chunk tier is specified, the entire channel is considered
to be a single chunk. If no global segment tier is specified, all {\em
  is\_init} and {\em is\_fin} are set to {\em no}.

Finally, if specified by the user, an arbitrary number of grouping
columns will be added to the tables that are derived from the
filenames. Their names are prefixed by \myRef{grp}. See the grouping
options \myRef{fsys:grp:*} in section \ref{sec:dtg} for details. Each
table file comes along with an R code template file with the same name and
the extension {\em .R} to read this table by the R software.

\subsection{Summary table files}
\label{sec:stf}

By setting \myRef{fsys:export:summary} to 1 the table output described
in section \ref{sec:otf} can be summarized per file and
analysis tier. Summarization for continuous-valued features is done in terms
of their mean, median, standard deviation, and inter-quartile
range. For categorical features as intonation contour classes the
unigram entropy is calculated. The resulting table is written to the
directory \myRef{fsys:export:dir} with the file stem
\myRef{fsys:export:stm} plus the suffix {\em summary} and the
extension {\em csv}. Columns are separated by a comma. There is one
row of statistic values per analysed tier in a file.  Each continous-valued
feature within each analysis tier is represented by four columns. For
features of the sets {\em glob} and {\em loc} for which there is only
one analysis tier the column names follow the pattern {\em
  featureSet\_featureName\_statisticMeasure}. The suffixes
representing the statistic measurements are listed in the table right
below. For features of all other sets with potentially more than one
analysis tier the column names are built like this: {\em
  featureSet\_analysisTierName\_featureName\_statisticMeasure}. Categorical
features are represented by one column each with the same name
building schema.

\begin{center}
  \begin{tabular}{r|l|l}
    suffix & meaning & feature type \\
    \hline
    m & arit. mean & continuous \\
    med & median & continuous \\
    sd & standard deviation & continuous \\
    iqr & inter-quartile range & continuous \\
    h & unigram entropy & categorical
  \end{tabular}
\end{center}

File level groupings, i.e. the {\em grp\_*} columns of the csv tables
decribed in section \ref{sec:otf}, are copied to the summary
table. File and channel index are given in the columns {\em fi} and
{\em ci}, respectively, the file stem is written to column {\em stm}.
Columns are sorted alphanumerically by their names.

Next to the csv file an R code template file is generated with the
same name and the extension {\em .R} to read the summary table by the
R software.

\subsection{Nested Python dictionary}
\label{sec:dict}

The pickle file which is outputted in \copac{config:fsys:export:dir}
contains a nested dictionary {\em copa} for the sake of further
processing within other Python projects.

On the top level {\em copa} can be subdivided into the sub-dictionaries
\begin{itemize}
\item \myRef{config}: configurations underlying the current analysis
  \item \myRef{export}: Pandas Dataframes of extracted features. One
    dataframe per feature set
\item \myRef{data}: extracted features in a nested dictionary described
  below
\item \myRef{clst}: contour clustering results
\item \myRef{val}: validation metrics for stylization and clustering
\end{itemize}

\subsubsection{Configuration sub-dictionary}
This sub-dictionary is accessed by \copac{copa['config']} and simply
contains a copy of the user-defined and default configurations which
are introduced in section \ref{sec:config}.

\subsubsection{Stylization feature table sub-dictionary}
Is stored in \copac{copa['export']}. The Pandas Dataframe for each
feature set can be accessed by the feature set's name. For example,
the local contour stylization parameters can be found here:

\begin{quote}
  \copac{copa['export']['loc']}
\end{quote}

Same for the other feature sets. All standard and rhythm features are
additionally accessible on the file level. Thus \copac{copa['export']}
is structured as follows\footnote{\copac{export:x} is to be expanded as \copac{copa['export'][x]}}:

\codat{export:bnd}{boundary features}{Pandas Dataframe}
\codat{export:gnl\_en}{standard energy features}{Pandas Dataframe}
\codat{export:gnl\_f0}{standard f0 features}{Pandas Dataframe}
\codat{export:gnl\_en\_file}{\ldots on file level}{Pandas Dataframe}
\codat{export:gnl\_f0\_file}{\ldots on file level}{Pandas Dataframe}
\codat{export:loc}{local f0 contour features}{Pandas Dataframe}
\codat{export:glob}{global f0 contour features}{Pandas Dataframe}
\codat{export:rhy\_en}{energy rhythm features}{Pandas Dataframe}
\codat{export:rhy\_f0}{f0 rhythm features}{Pandas Dataframe}
\codat{export:rhy\_en\_file}{\ldots on file level}{Pandas Dataframe}
\codat{export:rhy\_f0\_file}{\ldots on file level}{Pandas Dataframe}
\codat{export:voi}{voice quality features}{Pandas Dataframe}

\subsection{F0 files}

Three types of f0 tables can be exported:

\begin{itemize}
\item preprocessed f0
\item residual f0 (after removal of the global register component)
\item resynthesized f0 (superposition of global and local stylized
  component)
\end{itemize}

As the f0 table input format in each output table the first column
gives the time stamps, and the second till last columns contain the f0
values (in Hz) for the recording channels. The tables will be stored
below \myRef{fsys:export:dir} in sub-directories named after the type of
f0 output ({\em f0\_preproc, f0\_residual, f0\_resyn}). For each input
  f0 file an output file with the same name is generated.

\subsection{Log file}
The log file in {\em fsys:export:dir + fsys:export:stm + log.txt}
contains warnings, information about too short segments to be skipped,
and some validations below the line '\# validation':

\begin{center}
  \begin{tabular}{l|l}
    styl.glob.err\_prop & the percentage of global contour segments \\
 & where base and topline are crossing \\
styl.loc.rms\_mean & the mean RMSD between original and stylized \\
 & contour over all local contour segments
  \end{tabular}
\end{center}

The log file is not overwritten, but new logging information is
appended. Each session starts with the current time string in ISO 8601
format.

\section{Plotting}

To activate plotting, set

\begin{quote}
  \myRef{navigate:do\_plot=1}
\end{quote}

\paragraph{Browsing}

Browsing through stylizations can be carried out online (in order to
check for appropriate stylization parameter settings) or after
feature extraction, which is controlled by

\begin{quote}
  \myRef{plot:browse:time}
\end{quote}

To select the stylization to be plotted the corresponding branches in

\begin{quote}
  \myRef{plot:browse:typ:*:*}
\end{quote}

need to be set to true. E.g. \myRef{plot:browse:typ:complex:superpos=1}
produces plots as in Figure \ref{fig:superpos}.

It is possible to plot stylizations of segments with certain grouping
values only. This is achieved by specifying one or more

\begin{quote}
  \myRef{plot:browse:grp:myGroupingVariable:myGroupingValue}
\end{quote}

{\em myGroupingVariable} should match one of the strings in
\myRef{fsys:grp:lab}. The string {\em myGroupingValue} should match
one of the values assigned to {\em myGroupingVariable} by file name
parsing (cf section \ref{sec:dtg}).

\paragraph{Grouping}
One can also plot stylizations based on parameter centroids for a
specified grouping. By

\begin{quote}
  \myRef{plot:grp:typ:*:*=1}
\end{quote}

the user selects the stylization to be plotted. The grouping is defined by 

\begin{quote}
  \myRef{plot:grp:grouping}
\end{quote}

The entries in this list can be {\em lab} for item labels or the
grouping factor names specified in \myRef{fsys:grp:lab}. Centroids
will be plotted for each factor level combination.

Browsing and grouping plots can be saved as .png files by

\begin{quote}
  \myRef{plot:browse:save=1} \\
  \myRef{plot:grp:save=1}
\end{quote}

The {\em browse} mode output file names are the concatenation of
\myRef{fsys:pic:dir} + \myRef{fsys:pic:stm} + {\em final$|$online} +
      {\em typ} + {\em set} + {\em fileIndex} + {\em channelIndex} +
      {\em tierName} + {\em itemIndex}. {\em typ} and {\em set} refer
      to the *-keys in plot:browse:typ:*:* set to true.

The {\em grouping} mode output file names are concatenated from
\myRef{fsys:pic:dir} + \myRef{fsys:pic:stm} + {\em
  factorLevelCombination}.  One file is generated for each factor level
combination.

\newpage

\addcontentsline{toc}{section}{References}
\bibliographystyle{gerabbrv} \bibliography{copasul_manual_latest}

\end{document}